\documentclass[runningheads]{llncs}

\usepackage{eccv}

\UseRawInputEncoding

\usepackage{eccvabbrv}

\usepackage{graphicx}
\usepackage{booktabs}
\usepackage{multirow}
\usepackage{adjustbox}
\usepackage{moresize}
\usepackage[accsupp]{axessibility}  

\usepackage{hyperref}
\hypersetup{colorlinks}

\usepackage{orcidlink}

\usepackage{comment}

\usepackage[normalem]{ulem}
\newcommand{\stkout}[1]{\ifmmode\text{\sout{\ensuremath{#1}}}\else\sout{#1}\fi}

\begin{document}
\title{PDiscoFormer: Relaxing Part Discovery Constraints with Vision Transformers} 

\titlerunning{PDiscoFormer: Part Discovery with Transformers}

\author{Ananthu Aniraj\inst{1}\orcidlink{0009-0003-4521-036X} \and
Cassio F. Dantas\inst{2}\orcidlink{0000-0002-1934-0625} \and
Dino Ienco\inst{2}\orcidlink{0000-0002-8736-3132} \and
Diego Marcos\inst{1}\orcidlink{0000-0001-5607-4445}}

\authorrunning{A.~Aniraj et al.}

\institute{Inria, Univ. Montpellier, LIRMM, UMR TETIS, Montpellier, France
\email{\{ananthu.aniraj, diego.marcos\}@inria.fr} \and
Inria, Inrae, Univ. Montpellier, UMR TETIS, Montpellier, France
\email{\{cassio.fraga-dantas, dino.ienco\}@inrae.fr}
}
\maketitle

\begin{abstract}
  Computer vision methods that explicitly detect object parts and reason on them are a step towards inherently interpretable models. Existing approaches that perform part discovery driven by a fine-grained classification task make very restrictive assumptions on the geometric properties of the discovered parts; they should be small and compact.
  Although this prior is useful in some cases, in this paper we show that pre-trained transformer-based vision models, such as self-supervised DINOv2 ViT, enable the relaxation of these constraints.
  In particular, we find that a total variation (TV) prior, which allows for multiple connected components of any size, substantially outperforms previous work.
  We test our approach on three fine-grained classification benchmarks: CUB, PartImageNet and Oxford Flowers, and compare our results to previously published methods as well as a re-implementation of the state-of-the-art method PDiscoNet with a transformer-based backbone.
  We consistently obtain substantial improvements across the board, both on part discovery metrics and the downstream classification task, showing that the strong inductive biases in self-supervised ViT models require to rethink the geometric priors that can be used for unsupervised part discovery. Training code and pre-trained models are available at \url{https://github.com/ananthu-aniraj/pdiscoformer}.
\end{abstract}

\section{Introduction}
\label{sec:intro}

\begin{figure}[tb]
  \centering
  \includegraphics[width=\textwidth]{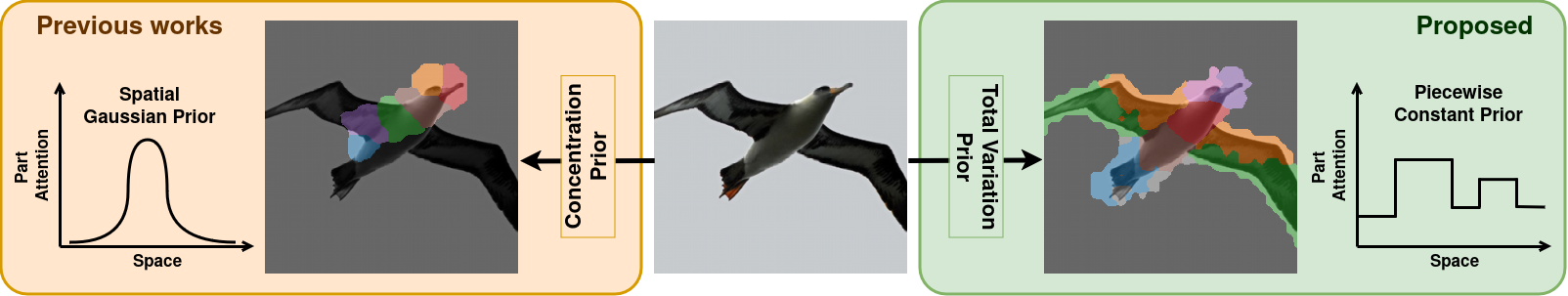}
  \caption{Concentration \emph{vs} total variation as part priors.}
  \label{fig:splash-fig}
\end{figure}

Deep neural networks, first in the form of convolutional neural networks (CNN) and, more recently, vision transformers (ViT), have revolutionized the field of computer vision. However, their black-box nature often limits their application in domains where interpretability is crucial. To ameliorate the understanding of the internal behaviour of deep learning models trained for computer vision tasks, post-hoc interpretability methods~\cite{lime, 8237336, ramaswamy2020ablation} are commonly used. These methods generate saliency maps indicating regions in the image that contributed most to a classification decision. However, these are only approximations to the actual internal process of the model and may be misleading~\cite{rudin2019stop}.

Recent efforts have been directed towards developing interpretable-by-design deep neural networks~\cite{chen2019looks, nauta2023pipnet, van2023pdisconet, hung:CVPR:2019, Huang2020InterpretableGrouping, nauta2021looks}. These approaches aim to identify semantically interpretable features in an image and utilize this information for solving a downstream task. 
Primarily tailored for fine-grained image classification, these methods seek to emulate human reasoning by subdividing the final task, such as classifying flower species or dog breeds, into multiple explicit intermediate stages, such as identifying lower-level semantic features like body parts, which in turn are used to provide a classification score.

However, improving interpretability while maintaining performance remains a challenge. 
Some models, such as those proposed in~\cite{van2023pdisconet, hung:CVPR:2019, marcos2020whale,Huang2020InterpretableGrouping}, rely on objective functions that impose priors on the shape and area of object parts such as compactness-inducing part shaping losses that, in practice, result in spatial Gaussian priors of fixed variance on the part maps.
While effective for specific datasets, like the commonly used CUB~\cite{WelinderEtal2010} benchmark for bird species classification, where assumptions like parts occurring only once in an image and being compact are beneficial, these assumptions may not generalize well to other use cases. 
For instance, in plant species classification, the presence of multiple instances of object parts such as leaves or flowers, as well as irregularly-shaped parts like branches or stems, renders these assumptions inadequate.

As we show in this work, such strong part shape priors are required when using a CNN-based backbone pre-trained on ImageNet; in order to reach acceptable classification results, it is imperative to fine-tune the whole backbone. 
This means that the only inductive bias that is kept throughout the part discovery process is the one brought in by the convolutional nature of the CNN architecture.
A strong part shape prior is required in order to obtain consistent part maps in this setting, leading to all recent methods for part discovery in the context of fine-grained classification to use a concentration loss that makes part maps small and compact~\cite{van2023pdisconet, hung:CVPR:2019,Huang2020InterpretableGrouping}.
Here, we propose a method that addresses this issue by refraining from assuming any priors on part shape or area. The only prior we assume is that the part maps should tend to be piece-wise constant and thus be formed of one or more spatially connected components. 
Our model is capable of localizing up to a fixed number of semantically consistent parts of an object without relying on part annotations, using only the classification labels as supervision signals. 
The lack of assumptions about part shape and size makes our approach suitable to various fine-grained image classification tasks, as demonstrated by our empirical results, generalizing beyond cases where compactness is a good part prior.

The \textbf{main contribution} of this paper is to build upon~\cite{van2023pdisconet} and relax its constraining concentration-based part prior by a more flexible total variation prior.
We investigate which additional elements are required to compensate for the flexibility of the total variation loss and find that a combination of an entropy loss and Gumbel-Softmax on the part maps allows for consistent and high quality part maps.
Our approach makes better use of the representation of self-supervised pre-trained ViT models and allows to apply the model to tasks with diverse types of parts.

\section{Related Work}
\label{sec:rel_work}

\noindent \textbf{Unsupervised Part Discovery.} Machine learning methods designed for unsupervised part discovery aim to identify a small number of semantically consistent parts, typically around $K=8$, shared among classes in a dataset. These methods visualize all discovered parts in an image through per-part saliency maps, facilitating easier interpretation. However, as this task is unsupervised, it often relies on assumptions about the discovered parts, such as their distribution and shape. For instance, Huang \etal \cite{Huang2020InterpretableGrouping} assume that the presence of each part across a mini-batch follows a beta distribution, while \cite{van2023pdisconet} assumes that discovered parts are present at least once in a mini-batch. Additionally, assumptions are often made about the shape of the parts. Methods such as \cite{van2023pdisconet, hung:CVPR:2019, marcos2020whale, xu2022particul} impose a spatial Gaussian prior with a fixed variance on the part maps, resulting in compact parts appearing once in the image. Such assumptions on part shape can be highly restrictive and may not generalize well across different scenarios. To address this limitation, our proposed method refrains from assuming such priors on part shape and only imposes a spatial prior that discovered parts should form connected components.

\noindent \textbf{Part Prototype Networks.} Chen \etal \cite{chen2019looks} introduced the Prototypical Part Network (ProtoPNet), an inherently interpretable method featuring a fixed number of learned prototypes per class. These prototypes represent visual patterns or features learned by the network during training. The model computes the similarity between each learned prototype and image patches and uses a linear combination of these similarity scores for classification. While explanations provided by ProtoPNet visualize all prototypes along with their weighted similarity scores, the sheer number of prototypes, often proportional to the number of classes, can lead to redundancy and interpretability challenges~\cite{nauta2021looks, gautam2023looks}. Subsequent works in this line of research have sought to address these challenges. Some approaches focus on reducing the number of prototypes shared per class~\cite{wang2021interpretable, rymarczyk2021protopshare}, while others explore sharing prototypes between classes~\cite{nauta2023pipnet, Nauta_2021_CVPR}. Despite these efforts, recent works like~\cite{nauta2023pipnet} still utilize a large number ($K > 500$) of prototypes, which can pose significant challenges for end-user interpretation.

\noindent \textbf{Regularization in Image Segmentation.} Early works on image segmentation made direct use of low-level image features, chiefly color, along with geometric priors, in order to find coherent segments. 
The Mumford-Shah model~\cite{mumford1989optimal}, originally devised for piece-wise image reconstruction, was applied to find segments that are as homogeneous as possible while trying to minimize the length of their boundary as a regularization~\cite{chan2001active}.
Total variation (TV)~\cite{rudin1992nonlinear} has been shown to be a powerful and efficient regularization to approximate this boundary length prior, and has been extensively used for piece-wise signal and image reconstruction~\cite{dobson1996recovery,strong2003edge}, as well as unsupervised image segmentation~\cite{unger2008tvseg,donoser2009saliency}.
With the advent of deep learning, TV regularization has proven valuable for style transfer~\cite{johnson2016perceptual} and edge-preserving smoothing~\cite{yeh2022total}.
In this work we explore the use of TV regularization for weakly-supervised part discovery in order to alleviate the issues brought by previously used priors.

\noindent \textbf{Emergent grouping of ViT representations.} Some recent research~\cite{xu2022groupvit, ranasinghe2023perceptual} has studied the emergence of object segmentation by supervising with image-text pairs. These methods learn to group similar patches in a ViT and assign them to textual concepts. They aim to identify the pixels in an input image corresponding to a predefined text prompt, focusing on segmenting whole objects rather than decomposing them into parts, as our model does with part discovery, and are limited to concepts that are represented in the text annotations.

\begin{figure}[tb]
  \centering
  \includegraphics[width=\textwidth]{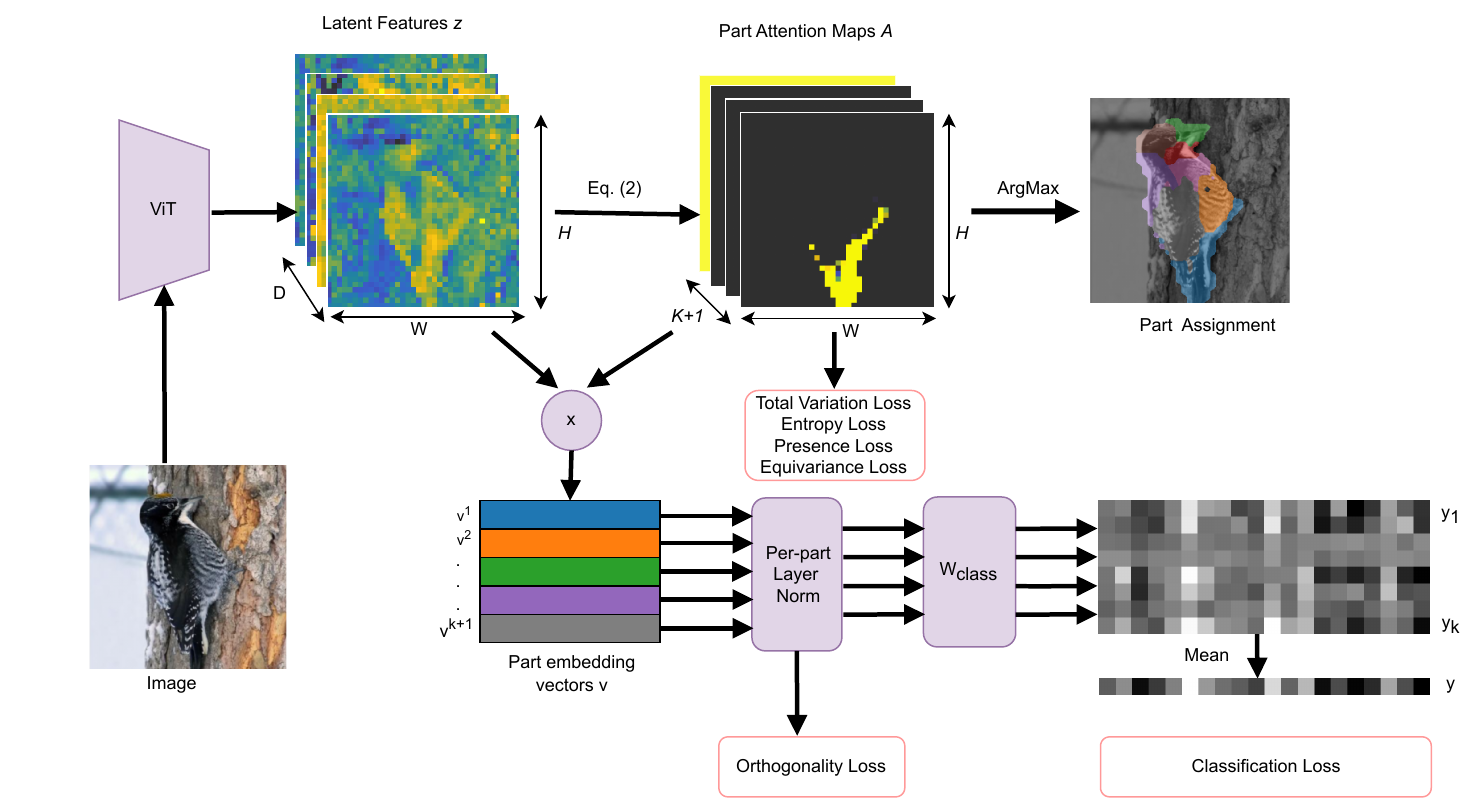}
  \caption{The \textbf{PDiscoFormer} architecture for part discovery. }
  \label{fig:pdisconet-main}
\end{figure}

\section{Methodology}
\label{sec:methodology}

Here we introduce an end-to-end trainable method (see~\cref{fig:pdisconet-main}) capable of automatically identifying $K$ discriminative parts (semantically interpretable image regions) useful for solving a fine-grained image classification task. 
In contrast to earlier methods~\cite{van2023pdisconet, Huang2020InterpretableGrouping}, our approach avoids any assumptions related to part shape, size or number of part instances. 
As in~\cite{van2023pdisconet, Huang2020InterpretableGrouping}, our model is weakly supervised and relies solely on image-level class labels, thus uncovering discriminative parts without requiring any additional supervision.
Our formulation is based on~\cite{van2023pdisconet}, from which we take take most elements of the architecture, except for the Gumbel-Softmax and the layer normalization-based part modulation. 
The total variation, entropy and background losses have not been used in previous part discovery works and are shown to be important in our results.

\subsection{Model architecture}
Let \( \mathbf{x} \in \mathbb{R}^{3 \times M \times N} \) be an image in the dataset, and let \( y \in \{1,2, ..., C\} \) represent the corresponding classification label.
Given a backbone model \( h_{\theta}(\cdot) \), we extract a feature tensor:
\begin{equation}
\mathbf{z} = h_{\theta}({\mathbf{x}}) \in \mathbb{R}^{D \times H \times W}
\end{equation}

We drop the class and register tokens of the ViT and reshape the patch tokens to obtain \( \mathbf{z} \). Following prior research~\cite{hung:CVPR:2019, Huang2020InterpretableGrouping, van2023pdisconet}, we compute attention maps \( \mathbf{A} \in [0, 1]^{ (K+1) \times H \times W} \), where the last channel in the first dimension is used to represent the background region in the image. 
These attention maps are computed using the negative squared Euclidean distance function between patch token features \( \mathbf{z}_{ij} \in \mathbb{R}^{D} \), \( i \in \{1, ..., H\} \), \( j \in \{1, ..., W\} \) and each of the learnable prototypes \( \mathbf{p}^{{k}} \in \mathbb{R}^{D} \), with $k \in \{1, \dots, K+1\} $, followed by a Gumbel-Softmax~\cite{jang2017categorical} across the \( {K}+1 \) channels:

\begin{equation}
\label{eq:gumbel}
a_{ij}^{{k}} = \frac{\exp\left(-\| \mathbf{z}_{ij} - \mathbf{p}^{{k}}\|^2 + \gamma_{k} \right)}{\sum_{{l=1}}^{{K+1}} \exp\left(-\| \mathbf{z}_{ij} - \mathbf{p}^{{l}}\|^2 + \gamma_{l}\right)}
\end{equation}
where $\gamma_{k}$ and  $\gamma_{l}$ are independently drawn samples from the \texttt{Gumbel(0,1)} distribution, and $a_{ij}^k$ represents an element of $\mathbf{A}$.

The attention maps are used to compute part embedding vectors \( \mathbf{v}^{{k}} \in \mathbb{R}^{D} \) by using the attention values to calculate a weighted average over the feature tensor \( \mathbf{z} \):

\begin{equation}
    \mathbf{v}^{{k}} = \frac{\sum_{i}\sum_{j} a_{ij}^{{k}} \mathbf{z}_{ij}}{HW}
\end{equation}

These embedding vectors are then modulated using a layer normalization \cite{ba2016layer} operation, with normalization applied over the feature and prototype dimensions. This operation helps to make each embedding vector distinct for the classifier:

\begin{equation}
\mathbf{v}_{m}^{k} = \frac{(\mathbf{v}^{k} - \text{E}[\mathbf{v}^{k}])}{\sqrt{\text{Var}[\mathbf{v}^{k}] + \epsilon}} \odot \mathbf{w}^k_{m} + \mathbf{b}^k_{m}
\end{equation}
with $\mathbf{w}^k_{m}, \mathbf{b}^k_{m} \in \mathbb{R}^{D}$ being the modulation weights and biases per part.
These modulated embedding vectors \( \mathbf{v}_{m}^{k} \) for $k\in\{1,\dots, K\}$ (we omit the background embedding) are then linearly projected to obtain a vector of class scores conditioned on the part embedding:
\begin{equation}
    \mathbf{y}^{k} = W_{c} \cdot \mathbf{v}_{m}^{k}    
\end{equation}

The same linear classifier $W_{c} \in \mathbb{R}^{C \times D}$ is applied to each of the modulated embedding vectors, i.e., for each of the $K$ discovered foreground parts. We then take the mean of the per-part class scores to obtain the classification scores:

\begin{equation}
    \mathbf{y} = \frac{1}{K} \sum_{k=1}^{K} \mathbf{y}^{k}
\end{equation}
Finally, a softmax function is employed to obtain the classification probabilities.

\noindent \textbf{Part dropout.}
\label{subsec:part_dropout}
We adopt dropout \cite{srivastava2014dropout} on the modulated embedding vectors $\mathbf{v}_{m}^{k}$, as proposed in~\cite{van2023pdisconet}, during model training. This drops entire part embedding vectors at a time and ensures that all discovered parts are discriminative, since discovering a single discriminative part would not be enough to solve the downstream classification task. 

\subsection{Loss Functions}
\label{subsec:loss_fns}

Our main contribution consists in proposing a set of part-shaping loss functions that are able to make the most out of the inductive biases learned by self-supervised ViT models.

\noindent \textbf{Classification Loss ($\mathcal{L}_{\text{c}}$).} Our model is mainly driven by the fine-grained classification objective during training. We use the standard negative log likelihood or multi-class cross-entropy for this purpose. 
While the classification loss helps ensure the discriminative nature of the discovered parts, it may not directly enforce the learned attention maps to represent salient parts of the object in the image. To address this, we incorporate additional objective functions to enforce priors on the discovered parts. These objectives effectively guide the learning process to focus on relevant regions of the image.

\noindent \textbf{Orthogonality Loss ($\mathcal{L}_{\perp}$).} We aim to discover parts that are semantically distinct. To achieve this, we apply the orthogonality loss, which encourages the learned part embedding vectors to be decorrelated from one another. This is accomplished by minimizing the cosine similarity between the embedding vectors. We apply this loss on the modulated embedding vectors \( \mathbf{v}_{m}^{k} \).
\begin{equation}
\mathcal{L}_{\perp}= \sum_{k=1}^{K+1} \sum_{l\neq k} \frac{\mathbf{v}_{m}^{k}\cdot \mathbf{v}_{m}^{l}}{\left \| \mathbf{v}_{m}^{k}\  \right \|\cdot \left \| \mathbf{v}_{m}^{l} \right \|}
\end{equation}

\noindent \textbf{Equivariance Loss ($\mathcal{L}_{\text{eq}}$).} We aim to detect the same parts even if the image is translated, rotated, or scaled. To achieve this, we formulate the equivariance loss, which encourages the learned part attention maps to be equivariant to rigid transformations. Specifically, given an input image, we use an affine transform $T$ with parameters randomly sampled from a pre-defined range and pass both the original image and the transformed image through our model. Then, we invert the transformation on the part attention maps of the transformed image. Finally, the equivariance loss is formulated as the cosine distance between the part attention maps of the original image and those of the corresponding transformed image.

Let $A^{k}(\mathbf{x})$ be a function that returns the $k^{th}$ attention map given an input image $\mathbf{x}$. Then:
\begin{equation}
\mathcal{L}_{\text{eq}} = 1 - \frac{1}{K} \sum_{k} \frac{\left \| A^{k}(\mathbf{x}) \cdot T^{-1}\left( A^{k}\left( T(\mathbf{x} \right) \right) \right \|}{\left \| A^{k}(\mathbf{x}) \right \| \cdot \left \| A^{k}\left( T(\mathbf{x}) \right) \right \|}
\end{equation}

\noindent \textbf{Presence Loss ($\mathcal{L}_{\text{p}}$).}
We assume that all discovered foreground parts are present in at least some images of the training dataset, while the background, the $(K+1)^{th}$ part, is expected to be present in all images in the dataset. To enforce this, we formulate the presence loss, which consists of two parts.
Firstly, we define a loss applied only to the discovered foreground parts, ensuring that each part is present at least once in a mini-batch. Given a batch $\{ \mathbf{x}_{1}, ..., \mathbf{x}_{B} \}$, the presence loss is calculated as:
\begin{equation}
 \mathcal{L}_{\text{p}_{1}} = 1 - \frac{1}{K} \sum_{k}\max_{b,i,j} \bar{a}^k_{ij}(\mathbf{x}_b)
\end{equation}
where $\bar{A^k}(\mathbf{x}_b) = \text{avgpool}(A^{k}(\mathbf{x}_b))$ is the output of 
a 2D average pooling operator with a small kernel size employed to prevent single-pixel attention maps.

Secondly, to ensure the presence of the background, we introduce a stricter presence loss, which is satisfied if the background is present in all images of the mini-batch. 
This loss is formulated as follows:
\begin{equation}
\mathcal{L}_{\text{p}_{0}}= -\frac{1}{B}\sum_{b} \log\left (
\max_{i,j} ~ m_{ij}  \bar{a}^{K+1}_{ij}(\mathbf{x}_b)
\right) 
\end{equation}
where $M = [m_{ij}]^{H \times W}$ is a soft mask  with $m_{ij} \in [0,1]$ that privileges entries placed farther from the image center (towards the borders):
\begin{equation}
m_{ij} = 2 \left( \frac{i-1}{H-1} - \frac{1}{2}\right)^2 + 2 \left( \frac{j-1}{W-1} - \frac{1}{2}\right)^2
\end{equation}

This loss captures the notion that the background is at the same time expected in every image and is more likely to appear near the boundaries of the image.

\noindent \textbf{Entropy Loss ($\mathcal{L}_{\text{ent}}$).}
We aim to ensure that each patch token is assigned to a unique part. To achieve this, we formulate an entropy~\cite{shannon1948mathematical} loss applied to the part attention maps:
\begin{equation}
\label{eq:entropy_loss}
\mathcal{L}_{\text{ent}} = \frac{-1}{K+1}\sum_{k=1}^{K+1} \sum_{ij}a^{k}_{ij} \log\left (a^{k}_{ij}  \right ) 
\end{equation}

\noindent \textbf{Total Variation Loss ($\mathcal{L}_{\text{tv}}$).}
We do not want to make any assumptions about the part shape. However, we would still like the discovered parts to be composed of one or, at most, a few connected components. To achieve this, we make use of a total variation loss~\cite{rudin1992nonlinear} applied to the part attention maps:

\begin{equation}
\mathcal{L}_{\text{tv}} = \frac{1}{HW} \sum_{k=1}^{K+1}\sum_{ij} \left |  \nabla a_{ij}^k \right |
\end{equation}
where $\nabla a_{ij}^k$ is the spatial image gradient of part map $A^k$ in location $ij$.



\section{Experimental Setup}
\label{sec:experiments}

We aim to perform part discovery using only image-level class labels as supervision signals, assuming that the discovered parts are shared among classes, which is common in fine-grained image classification settings. Moreover, we aim to compare our model with similar methods in the literature in scenarios where they have been empirically shown to work well, as well as in new and more challenging scenarios where underlying data characteristics may require more flexible geometric modeling. 

All models were implemented in PyTorch. We used the DinoV2 \cite{oquab2023dinov2} ViT \cite{Dosovitskiy2021AnScale} model with register tokens \cite{darcet2023vision} as the starting weights for all experiments, utilizing the Base (ViT-B) variant. The training settings and compute requirements of our model are detailed in~\cref{sec-app:train_settings} and \cref{sec-app:compute_req} respectively.

\subsection{Datasets}
\label{subsec:datasets}

\noindent \textbf{CUB Dataset.} The CUB dataset \cite{WelinderEtal2010} consists of images depicting 200 bird species, with 5,994 images in the training set and 5,794 images in the testing set. 
Keypoint annotations for 15 different bird body parts are available for each image.
Bird parts are typically compact, although their presence may vary in images due to changes in pose and occlusion. Most images contain only a single instance of a bird.
For this dataset, we evaluate the results using keypoint regression (Kp) using ground truth part keypoint annotations to compare them with centroids discovered by the model. 
Furthermore, we compute the Normalized Mutual Information (NMI) and Adjusted Rand Index (ARI) to quantify the clustering quality of the discovered parts, as in related literature~\cite{Choudhury2021UnsupervisedReconstruction}.

\noindent \textbf{PartImageNet Dataset.} The PartImageNet dataset \cite{he2022partimagenet} comprises 158 classes distributed among 11 super-classes, such as airplanes, birds, and snakes. 
Each super-class is associated to between two and five possible parts, for a total of 41 part classes. 
Additionally, test images may contain multiple instances of objects, sometimes from different super-classes, while the classification label provides information about a single object. 
Notably, several super-classes, including snakes, airplanes, and reptiles, often exhibit irregular and non-compact part shapes.
For our experiments, we use the \texttt{PartImageNet OOD} variant. Although originally intended for out-of-distribution detection, this variant is commonly used in related literature. We train our models on 14,865 training images and test on 1,658 images, respectively.
For evaluation, we derive part centroids from ground truth part masks and calculate NMI and ARI scores to assess clustering quality. 
Additionally, we include an experiment on the newly released \texttt{PartImageNet Seg} variant in~\cref{sec-app:partimagenet-seg}.


\noindent \textbf{Oxford 102 Flower Dataset.} The Oxford 102 Flower Dataset~\cite{Nilsback08} comprises 8,189 images divided into 102 flower classes. The training and validation sets contain 10 images per class, totaling 1,020 images each, while the test set consists of 6,149 images, with a minimum of 20 images per class.
Since no part annotations are provided for this dataset, we evaluate the quality of the discovered parts by combining them and comparing the result to the provided foreground/background masks.
We evaluate model performance by calculating the mean Intersection over Union (mIoU) for the foreground mask.
This dataset often includes multiple instances of parts in a single image, such as flowers, as well as parts with irregular shapes, such as flower petals.

\section{Experiments}

\subsection{Comparison with State-of-the-Art}
\label{sec:exp1}

\begin{table}[t]
\centering
\caption{Quantitative comparison of the part discovery and classification capabilities of our model with related approaches from the literature. In this table, $K$ refers to the number of parts. The method with the label $^{**}$ does not use any class supervision.}
\label{tab:quant-results}
\begin{adjustbox}{width=\linewidth}
\begin{tabular}{c|c|cccc|c|ccc|c|cc}
 &
  \multicolumn{5}{c|}{CUB(\%)} &
  \multicolumn{4}{c|}{PartImageNet OOD(\%)} &
  \multicolumn{3}{c}{Flowers(\%)} \\ \cline{2-13} 
Method &
  $K$ &
  Kp~$\downarrow$ &
  NMI~$\uparrow$ &
  ARI~$\uparrow$ &
  \begin{tabular}[c]{@{}c@{}}Top-1 \\ Acc.~$\uparrow$\end{tabular} &
  $K$ &
  NMI~$\uparrow$ &
  ARI~$\uparrow$ &
  \begin{tabular}[c]{@{}c@{}}Top-1 \\ Acc.~$\uparrow$\end{tabular} &
  $K$ &
  \begin{tabular}[c]{@{}c@{}}Fg.\\ mIoU~$\uparrow$\end{tabular} &
  \begin{tabular}[c]{@{}c@{}}Top-1\\ Acc.~$\uparrow$\end{tabular} \\ \hline
\multirow{3}{*}{Dino~\cite{amir2021deep}$^{**}$} &
  4 &
  - &
  31.18 &
  11.21 &
  - &
  8 &
  19.17 &
  7.59 &
  - &
  2 &
  54.95 &
  - \\
 &
  8 &
  - &
  47.21 &
  19.76 &
  - &
  25 &
  31.46 &
  14.16 &
  - & 4 &
  55.11 &
  - \\
 &
  16 &
  - &
  50.57 &
  26.14 &
  - &
  50 &
  37.81 &
  16.50 &
  - &
  8 &
  54.44 &
  - \\ \hline
\multirow{3}{*}{Huang~\cite{Huang2020InterpretableGrouping}} &
  4 &
  11.51 &
  29.74 &
  14.04 &
  87.30 &
  8 &
  5.88 &
  1.53 &
  74.22 &
  2 &
  29.92 &
  93.07 \\
 &
  8 &
  11.60 &
  35.72 &
  15.90 &
  86.05 &
  25 &
  7.56 &
  1.25 &
  73.56 &
  4 &
  33.22 &
  93.14 \\
 &
  16 &
  12.60 &
  43.92 &
  21.10 &
  85.93 &
  50 &
  10.19 &
  1.05 &
  73.20 &
  8 &
  17.26 &
  92.86 \\ \hline
\multirow{3}{*}{PDiscoNet~\cite{van2023pdisconet}} &
  4 &
  9.12 &
  37.82 &
  15.26 &
  86.17 &
  8 &
  27.13 &
  8.76 &
  88.58 &
  2 &
  19.04 &
  77.51 \\
 &
  8 &
  8.52 &
  50.08 &
  26.96 &
  86.72 &
  25 &
  32.41 &
  10.69 &
  89.00 &
  4 &
  34.76 &
  83.05 \\
 &
  16 &
  7.67 &
  56.87 &
  38.05 &
  87.49 &
  50 &
  41.49 &
  14.17 &
  86.06 &
  8 &
  49.10 &
  81.04 \\ \hline
\multirow{3}{*}{\begin{tabular}[c]{@{}c@{}}PdiscoNet\\ + ViT-B~\cite{darcet2023vision}\end{tabular}} &
  4 &
  7.70 &
  52.59 &
  26.66 &
  88.61 &
  8 &
  19.28 &
  34.72 &
  90.95 &
  2 &
  4.92 &
  92.75 \\
 &
  8 &
  6.34 &
  65.01 &
  37.90 &
  86.95 &
  25 &
  28.23 &
  50.35 &
  90.29 &
  4 &
  1.95 &
  95.48 \\
 &
  16 &
  5.95 &
  68.63 &
  43.41 &
  84.04 &
  50 &
  29.48 &
  27.80 &
  89.69 &
  8 &
  13.18 &
  97.40 \\ \hline
\multirow{3}{*}{\begin{tabular}[c]{@{}c@{}}PDiscoFormer\\ (Ours)\end{tabular}} &
  4 &
  7.41 &
  58.13 &
  25.11 &
  \textbf{89.06} &
  8 &
  29.00 &
  52.40 &
  89.75 &
  2 &
  \textbf{73.62} &
  99.61 \\
 &
  8 &
  5.99 &
  69.87 &
  43.49 &
  88.79 &
  25 &
  44.71 &
  59.27 &
  90.77 &
  4 &
  73.32 &
  99.54 \\
 &
  16 &
  \textbf{5.74} &
  \textbf{73.38} &
  \textbf{55.83} &
  88.72 &
  50 &
  \textbf{46.29} &
  \textbf{62.21} &
  \textbf{91.01} &
  8 &
  69.59 &
  \textbf{99.64}
\end{tabular}
\end{adjustbox}
\end{table}

We conduct a quantitative analysis of our model's part discovery and classification capabilities and compare it with those of related methods from the literature. Additionally, we compare our model against a re-implementation of the state-of-the-art PDiscoNet~\cite{van2023pdisconet}, where we replaced the backbone with the variant used in this paper. The results of this comparison are summarized in~\cref{tab:quant-results}. 
\begin{figure}[t]
\centering
\setlength\tabcolsep{1.5pt} 
\begin{ssmall}
\centering
 \begin{tabular}{cccccccccc}
  \rotatebox[origin=tl]{90}{{\parbox{1.05cm}{\centering Image}}} &
 \includegraphics[width=0.1\textwidth]{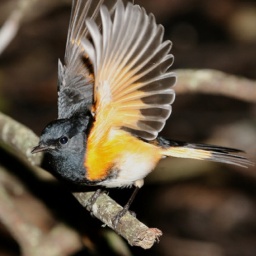} &
 \includegraphics[width=0.1\textwidth]{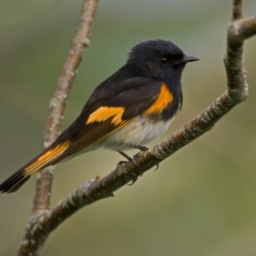}  &
 \includegraphics[width=0.1\textwidth]{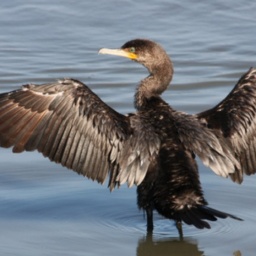}  &
 \includegraphics[width=0.1\textwidth]{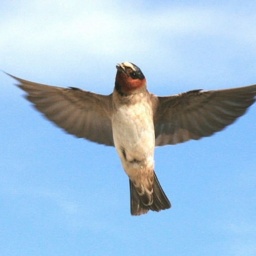}  &
 \includegraphics[width=0.1\textwidth]{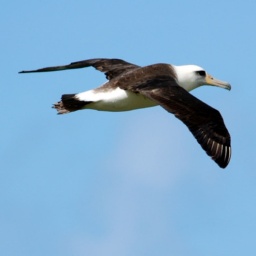}  &
 \includegraphics[width=0.1\textwidth]{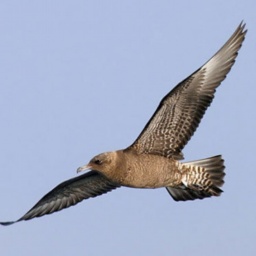}  &
 \includegraphics[width=0.1\textwidth]{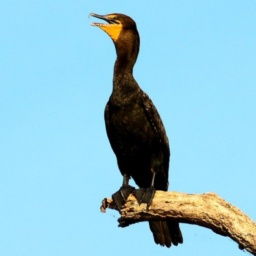}  &
 \includegraphics[width=0.1\textwidth]{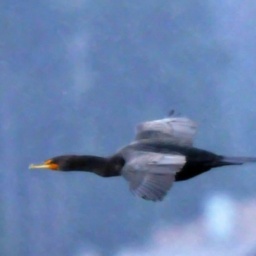}  &
 \includegraphics[width=0.1\textwidth]{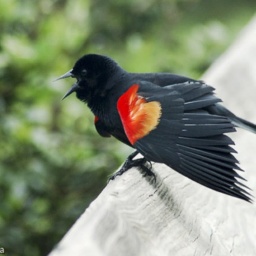}  \\
\rotatebox{90}{{\parbox{1.05cm}{\centering PDiscoNet}}} &
     \includegraphics[width=0.1\textwidth]{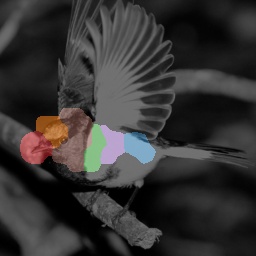} &
 \includegraphics[width=0.1\textwidth]{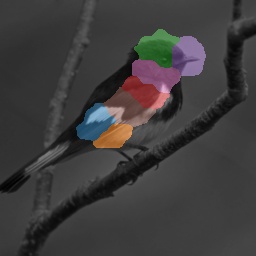}  &
 \includegraphics[width=0.1\textwidth]{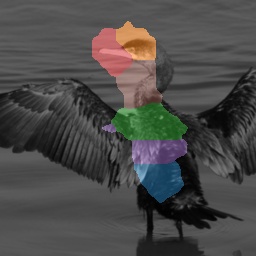}  &
 \includegraphics[width=0.1\textwidth]{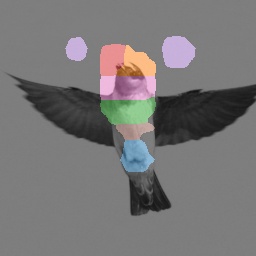}  &
 \includegraphics[width=0.1\textwidth]{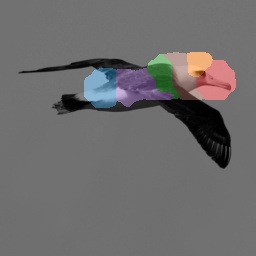}  &
 \includegraphics[width=0.1\textwidth]{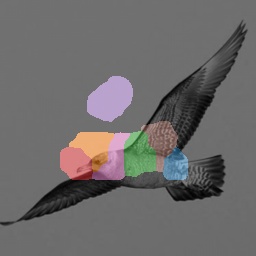}  &
 \includegraphics[width=0.1\textwidth]{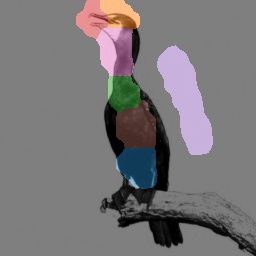}  &
 \includegraphics[width=0.1\textwidth]{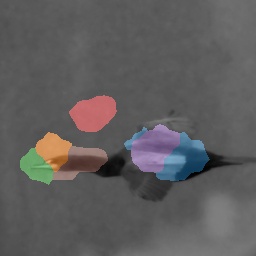}  &
 \includegraphics[width=0.1\textwidth]{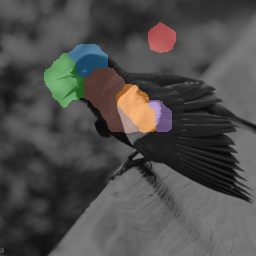} \\
 \rotatebox{90}{{\parbox{1.05cm}{\centering PDiscoNet + ViT-B}}} &
     \includegraphics[width=0.1\textwidth]{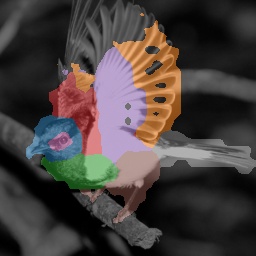} &
 \includegraphics[width=0.1\textwidth]{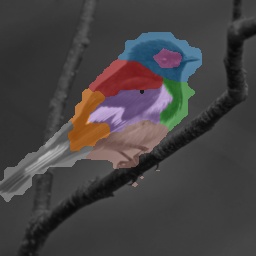}  &
 \includegraphics[width=0.1\textwidth]{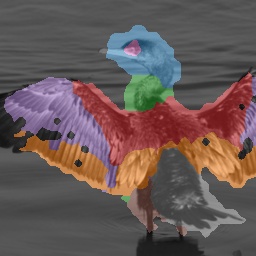}  &
 \includegraphics[width=0.1\textwidth]{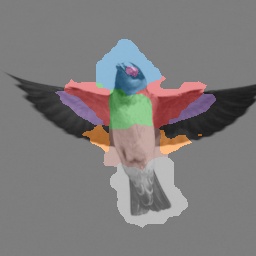}  &
 \includegraphics[width=0.1\textwidth]{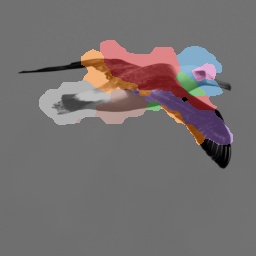}  &
 \includegraphics[width=0.1\textwidth]{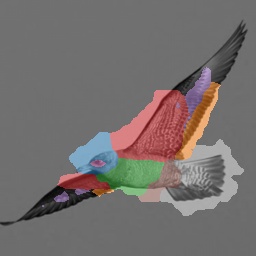}  &
 \includegraphics[width=0.1\textwidth]{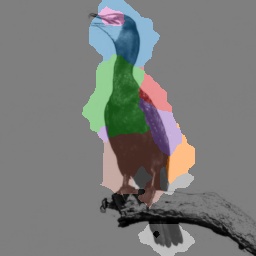}  &
 \includegraphics[width=0.1\textwidth]{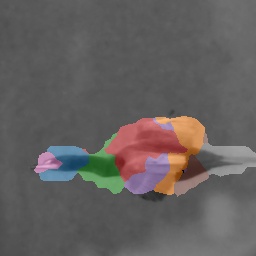}  &
 \includegraphics[width=0.1\textwidth]{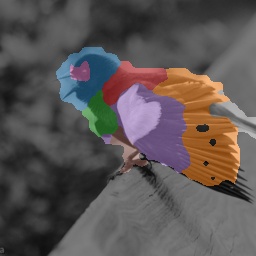} \\
\rotatebox[origin=tl]{90}{{\parbox{0.95cm}{\centering Ours}}} &
   \includegraphics[width=0.1\textwidth]{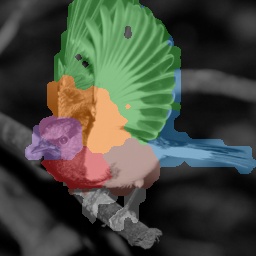} &
 \includegraphics[width=0.1\textwidth]{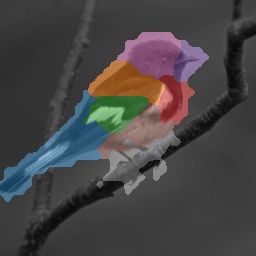}  &
 \includegraphics[width=0.1\textwidth]{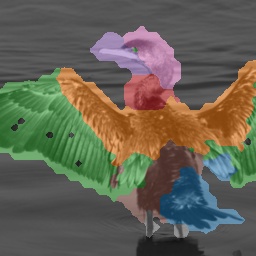}  &
 \includegraphics[width=0.1\textwidth]{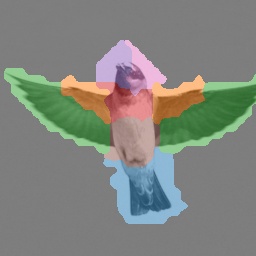}  &
 \includegraphics[width=0.1\textwidth]{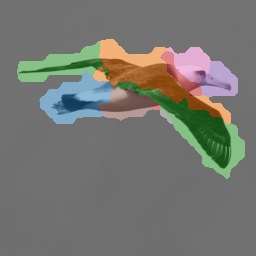}  &
 \includegraphics[width=0.1\textwidth]{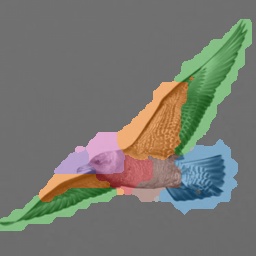}  &
 \includegraphics[width=0.1\textwidth]{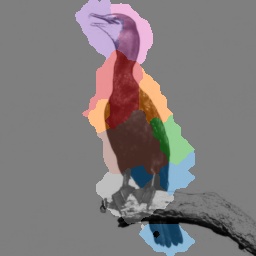}  &
 \includegraphics[width=0.1\textwidth]{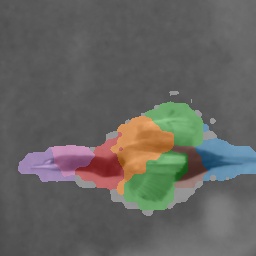}  &
 \includegraphics[width=0.1\textwidth]{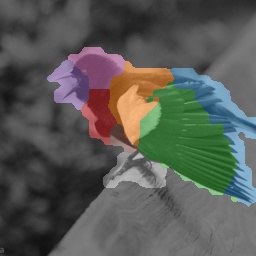} \\
\end{tabular}
\end{ssmall}
\caption{Qualitative results on CUB for $K=8$.}
\label{fig:qual-cub}
\centering
\setlength\tabcolsep{1.5pt} 
\begin{ssmall}
\centering
 \begin{tabular}{cccccccccc}
 \rotatebox[origin=tl]{90}{{\parbox{1.05cm}{\centering Image}}} &
 \includegraphics[width=0.1\textwidth]{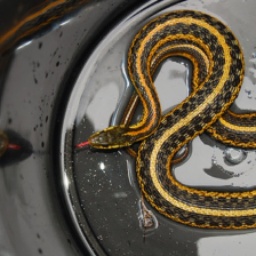} &
 \includegraphics[width=0.1\textwidth]{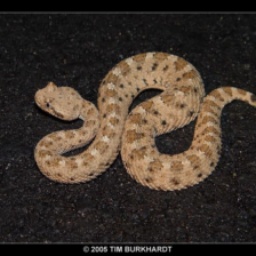}  &
 \includegraphics[width=0.1\textwidth]{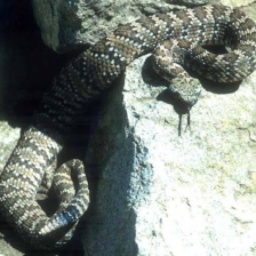}  &
 \includegraphics[width=0.1\textwidth]{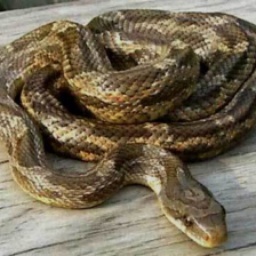}  &
 \includegraphics[width=0.1\textwidth]{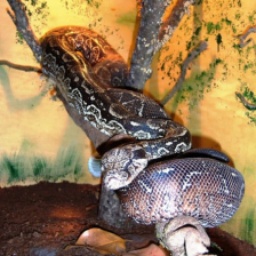}  &
 \includegraphics[width=0.1\textwidth]{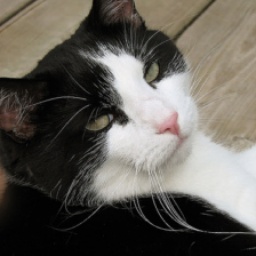}  &
 \includegraphics[width=0.1\textwidth]{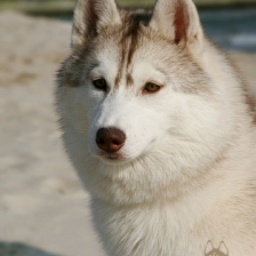}  &
 \includegraphics[width=0.1\textwidth]{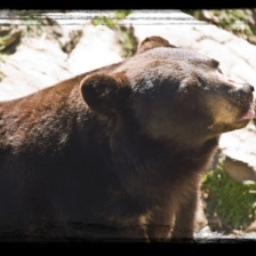}  &
 \includegraphics[width=0.1\textwidth]{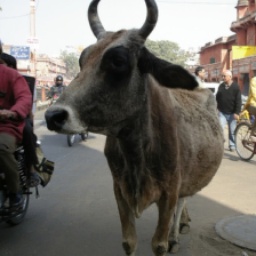}  \\
\rotatebox{90}{{\parbox{1.05cm}{\centering PDiscoNet}}} &
 \includegraphics[width=0.1\textwidth]{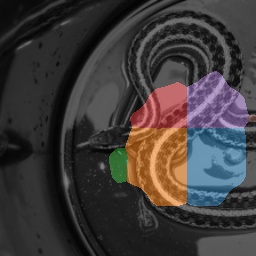} &
 \includegraphics[width=0.1\textwidth]{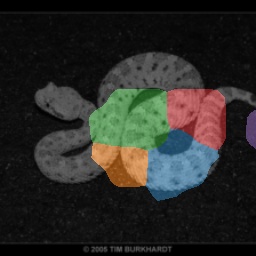}  &
 \includegraphics[width=0.1\textwidth]{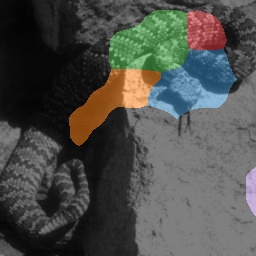}  &
 \includegraphics[width=0.1\textwidth]{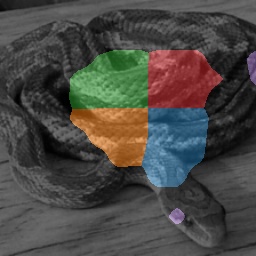}  &
 \includegraphics[width=0.1\textwidth]{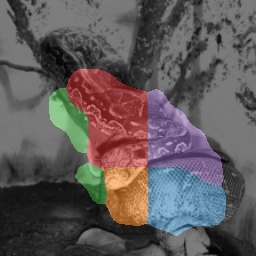}  &
 \includegraphics[width=0.1\textwidth]{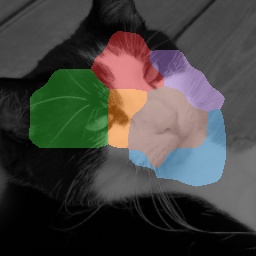}  &
 \includegraphics[width=0.1\textwidth]{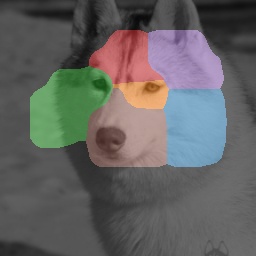}  &
 \includegraphics[width=0.1\textwidth]{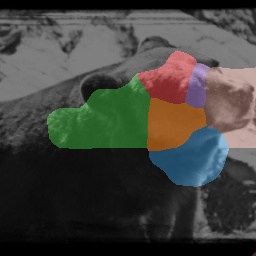}  &
 \includegraphics[width=0.1\textwidth]{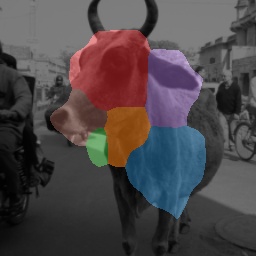} \\
 \rotatebox{90}{{\parbox{1.05cm}{\centering PDiscoNet + ViT-B}}} &
 \includegraphics[width=0.1\textwidth]{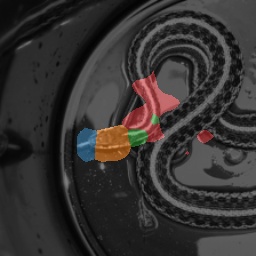} &
 \includegraphics[width=0.1\textwidth]{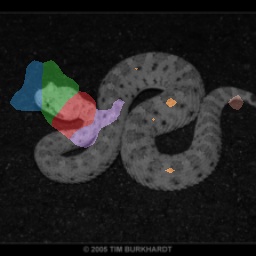}  &
 \includegraphics[width=0.1\textwidth]{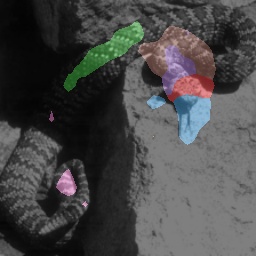}  &
 \includegraphics[width=0.1\textwidth]{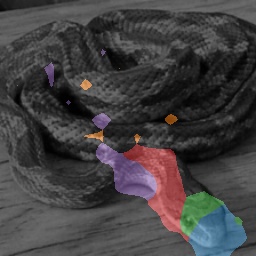}  &
 \includegraphics[width=0.1\textwidth]{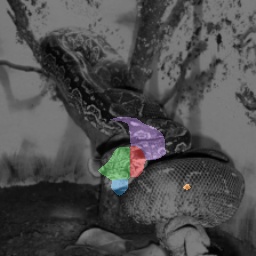}  &
 \includegraphics[width=0.1\textwidth]{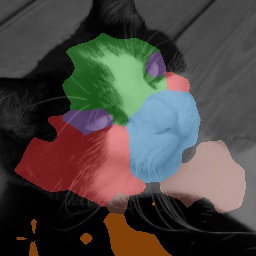}  &
 \includegraphics[width=0.1\textwidth]{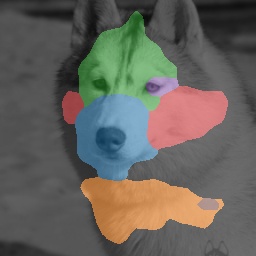}  &
 \includegraphics[width=0.1\textwidth]{images/qualitative_analysis/pdisconet_resnet/partimagenet/Quadruped/0_36_part_imagenet_ood.jpg}  &
 \includegraphics[width=0.1\textwidth]{images/qualitative_analysis/pdisconet_resnet/partimagenet/Quadruped/0_41_part_imagenet_ood.jpg} \\
\rotatebox[origin=tl]{90}{{\parbox{0.95cm}{\centering Ours}}} &
\includegraphics[width=0.1\textwidth]{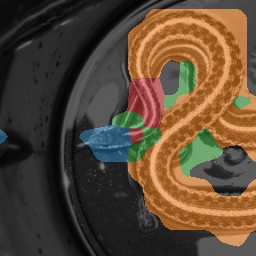} &
 \includegraphics[width=0.1\textwidth]{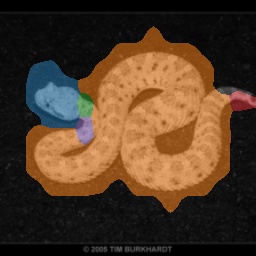}  &
 \includegraphics[width=0.1\textwidth]{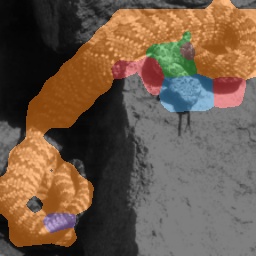}  &
 \includegraphics[width=0.1\textwidth]{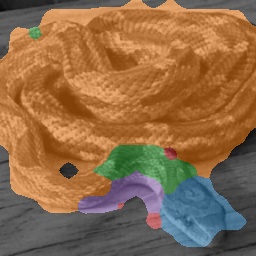}  &
 \includegraphics[width=0.1\textwidth]{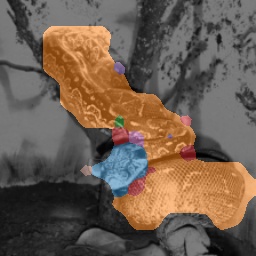}  &
 \includegraphics[width=0.1\textwidth]{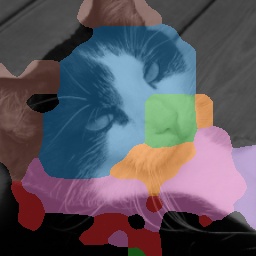}  &
 \includegraphics[width=0.1\textwidth]{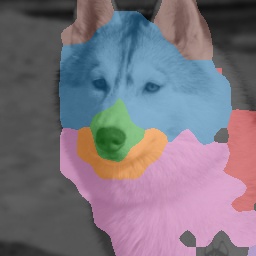}  &
 \includegraphics[width=0.1\textwidth]{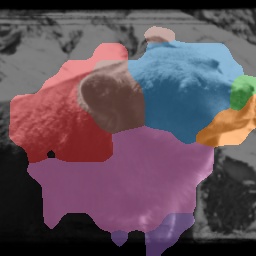}  &
 \includegraphics[width=0.1\textwidth]{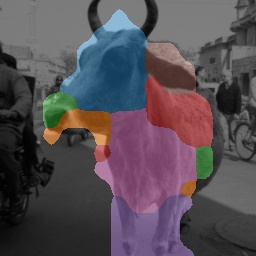} \\
\end{tabular}
\end{ssmall}
\caption{Qualitative results on PartImageNet for $K=8$.}
\label{fig:qual-pimagenet}
\end{figure}
From the results, our method consistently outperforms existing approaches for both part discovery and image classification across all datasets. We attribute this to the synergistic effects of the strong inductive biases ingrained in the self-supervised ViT backbone and the enhanced flexibility of the proposed combination of geometric priors. Particularly noteworthy are the significant performance gains observed on the Flowers and PartImageNet datasets, which contain objects with irregular shapes occurring in multiple instances within the same image, underscoring the importance of our geometric part prior that favors piece-wise constant and spatially connected components.
Even on CUB, where most parts fit well into the compactness prior, we see a consistent improvements of results with respect to PDiscoNet, even when using the same ViT backbone.
With 16 parts, our model achieves 55.8\% ARI and 73.4\% NMI, compared to 43.4\% ARI and 68.6\% NMI of the best competing method, PDiscoNet with a ViT backbone.
Despite competitive performance in classification tasks on CUB and PartImageNet, the PdiscoNet+ViT-B model tends to degrade in accuracy as part complexity increases, particularly in CUB, unlike our model, which maintains robust accuracy for any number of parts.
In the Flowers dataset, all competing methods result in subpar performance, while our approach excels at both detecting the foreground and classification no matter the number of parts. In this dataset, PDiscoNet+ViT-B results in the second best classification accuracy, 97.4\%, after the 99.6\% obtained by our method, but completely fails to locate the foreground, with an mIoU of 13.2\%, against 73.6\% for PDiscoFormer. Additionally, we extend our comparisons by applying our proposed losses to a ResNet backbone and a fully frozen ViT architecture, showing that a pre-trained ViT is required to fully profit from the proposed set of losses and that fine-tuning the position, class and register tokens, while keeping the rest of the ViT frozen, results in better performances than leaving the whole ViT frozen. The results of these additional comparisons are presented in~\cref{sec-app:backbone}.

\begin{figure}[t]
\centering
\setlength\tabcolsep{1.5pt} 
\begin{ssmall}
\centering
 \begin{tabular}{cccccccccc}
 \rotatebox[origin=tl]{90}{{\parbox{1.05cm}{\centering Image}}} &
 \includegraphics[width=0.1\textwidth]{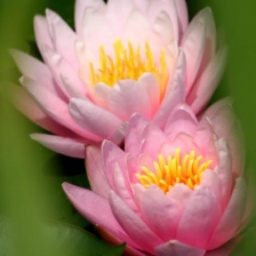} &
 \includegraphics[width=0.1\textwidth]{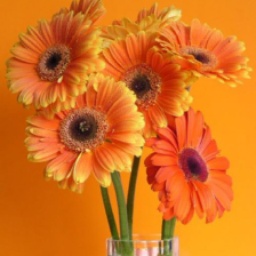}  &
 \includegraphics[width=0.1\textwidth]{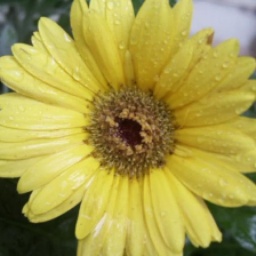}  &
 \includegraphics[width=0.1\textwidth]{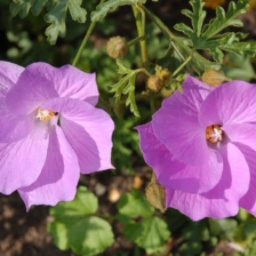}  &
 \includegraphics[width=0.1\textwidth]{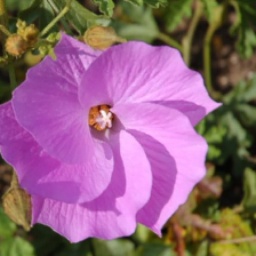}  &
 \includegraphics[width=0.1\textwidth]{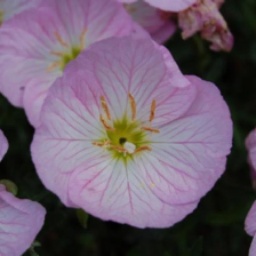}  &
 \includegraphics[width=0.1\textwidth]{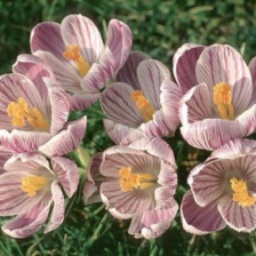}  &
 \includegraphics[width=0.1\textwidth]{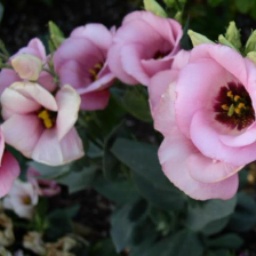}  &
 \includegraphics[width=0.1\textwidth]{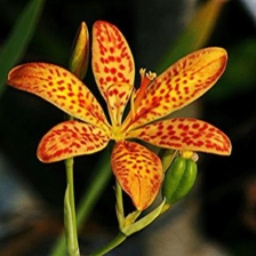}  \\
\rotatebox{90}{{\parbox{1.05cm}{\centering PDiscoNet}}} &
 \includegraphics[width=0.1\textwidth]{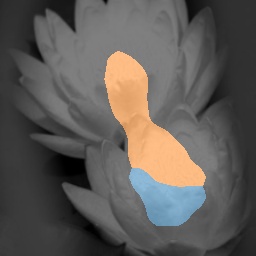} &
 \includegraphics[width=0.1\textwidth]{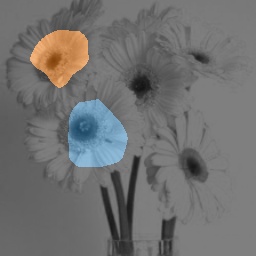}  &
 \includegraphics[width=0.1\textwidth]{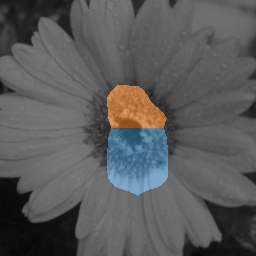}  &
 \includegraphics[width=0.1\textwidth]{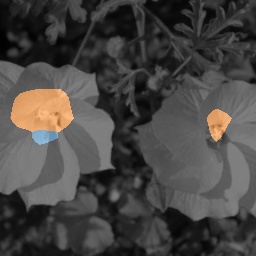}  &
 \includegraphics[width=0.1\textwidth]{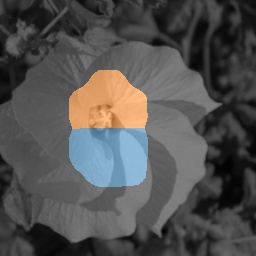}  &
 \includegraphics[width=0.1\textwidth]{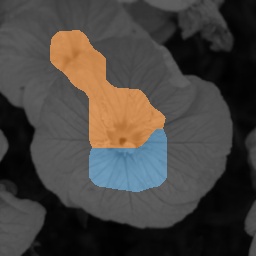}  &
 \includegraphics[width=0.1\textwidth]{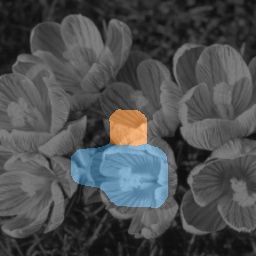}  &
 \includegraphics[width=0.1\textwidth]{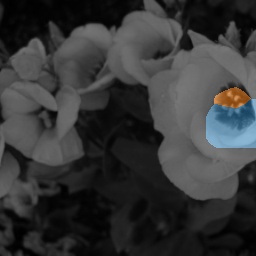}  &
 \includegraphics[width=0.1\textwidth]{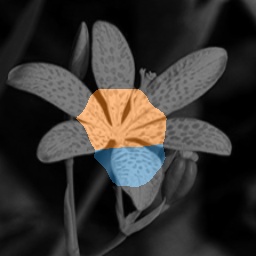} \\
 \rotatebox{90}{{\parbox{1.05cm}{\centering PDiscoNet + ViT-B}}} &
 \includegraphics[width=0.1\textwidth]{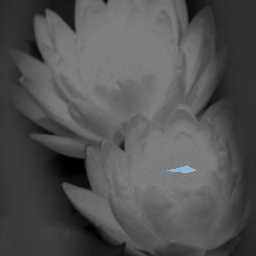} &
 \includegraphics[width=0.1\textwidth]{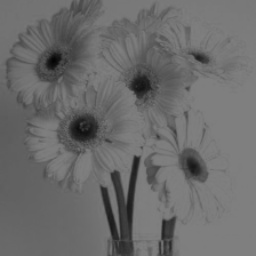}  &
 \includegraphics[width=0.1\textwidth]{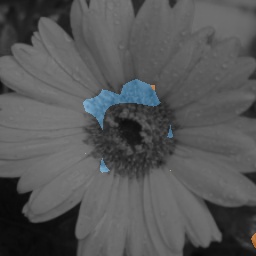}  &
 \includegraphics[width=0.1\textwidth]{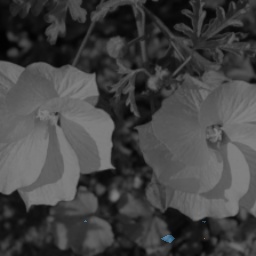}  &
 \includegraphics[width=0.1\textwidth]{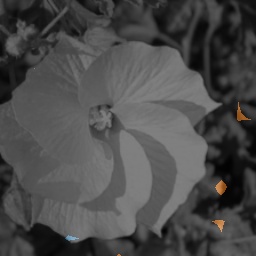}  &
 \includegraphics[width=0.1\textwidth]{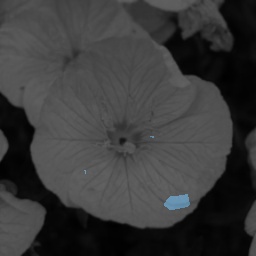}  &
 \includegraphics[width=0.1\textwidth]{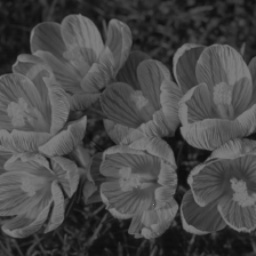}  &
 \includegraphics[width=0.1\textwidth]{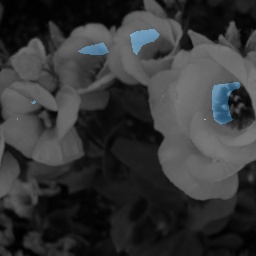}  &
 \includegraphics[width=0.1\textwidth]{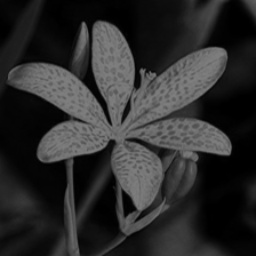} \\
\rotatebox[origin=tl]{90}{{\parbox{0.95cm}{\centering Ours}}} &
\includegraphics[width=0.1\textwidth]{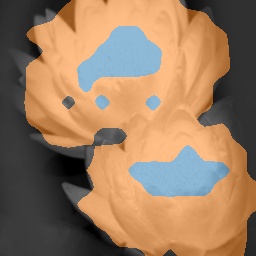} &
 \includegraphics[width=0.1\textwidth]{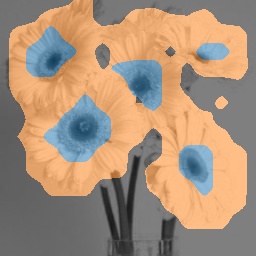}  &
 \includegraphics[width=0.1\textwidth]{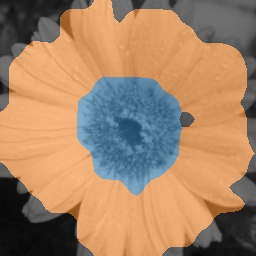}  &
 \includegraphics[width=0.1\textwidth]{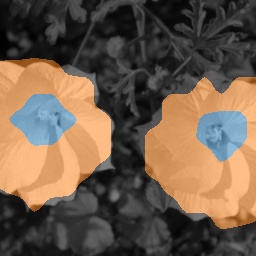}  &
 \includegraphics[width=0.1\textwidth]{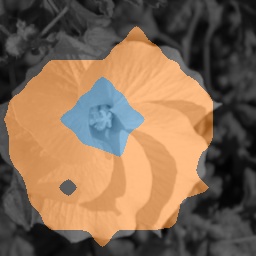}  &
 \includegraphics[width=0.1\textwidth]{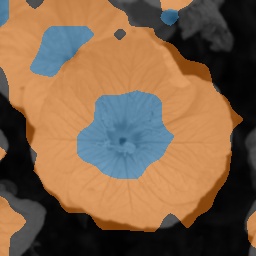}  &
 \includegraphics[width=0.1\textwidth]{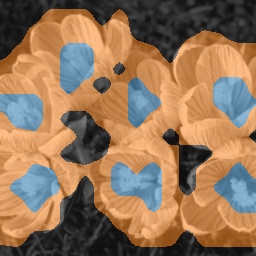}  &
 \includegraphics[width=0.1\textwidth]{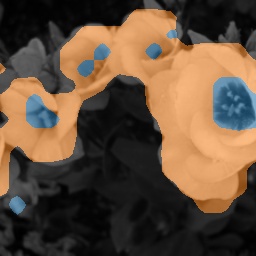}  &
 \includegraphics[width=0.1\textwidth]{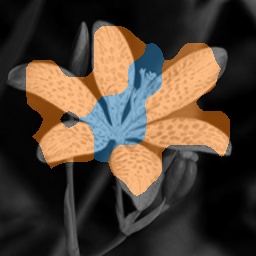} \\
 \end{tabular}
\end{ssmall}  
\caption{Qualitative results on Flowers for $K=2$.}
\label{fig:qual-flowers}
\centering
\setlength\tabcolsep{1.5pt} 
\begin{ssmall}
\centering
 \begin{tabular}{cccccccccc}
 \rotatebox[origin=tl]{90}{{\parbox{0.95cm}{\centering Image}}} &
 \includegraphics[width=0.1\textwidth]{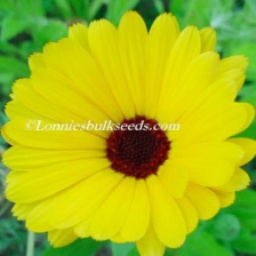} &
 \includegraphics[width=0.1\textwidth]{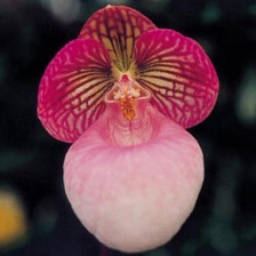}  &
 \includegraphics[width=0.1\textwidth]{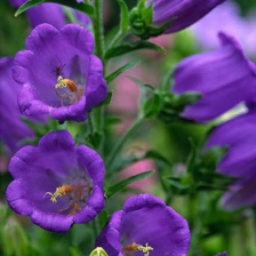}  &
 \includegraphics[width=0.1\textwidth]{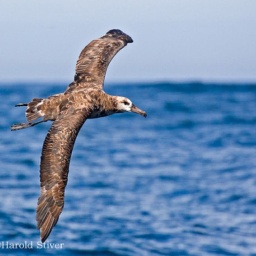}  &
 \includegraphics[width=0.1\textwidth]{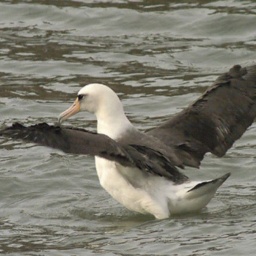}  &
 \includegraphics[width=0.1\textwidth]{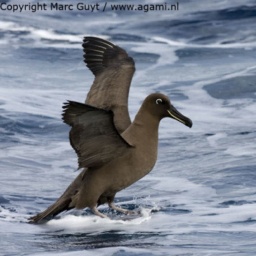}  &
 \includegraphics[width=0.1\textwidth]{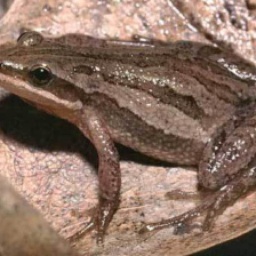}  &
 \includegraphics[width=0.1\textwidth]{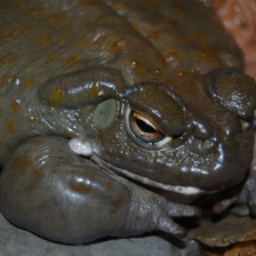}  &
 \includegraphics[width=0.1\textwidth]{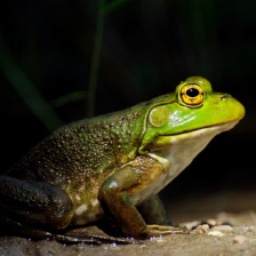}  \\
\rotatebox[origin=tl]{90}{{\parbox{0.95cm}{\centering K\\Low}}} &
 \includegraphics[width=0.1\textwidth]{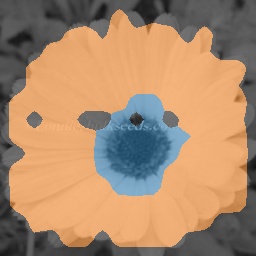} &
 \includegraphics[width=0.1\textwidth]{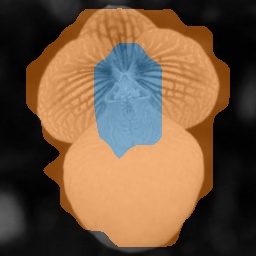}  &
 \includegraphics[width=0.1\textwidth]{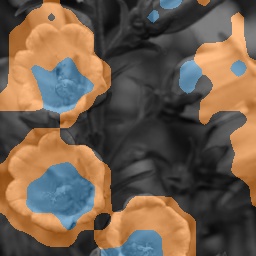}  &
 \includegraphics[width=0.1\textwidth]{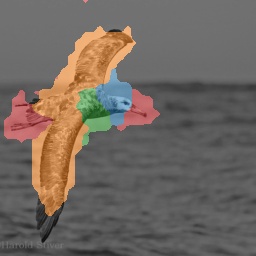}  &
 \includegraphics[width=0.1\textwidth]{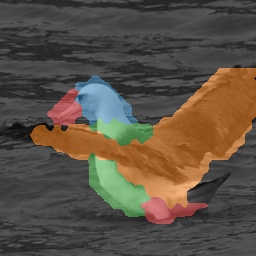}  &
 \includegraphics[width=0.1\textwidth]{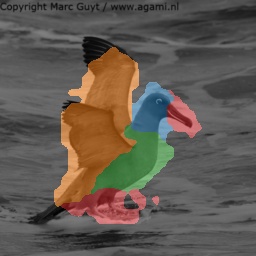}  &
 \includegraphics[width=0.1\textwidth]{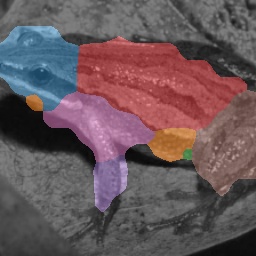}  &
 \includegraphics[width=0.1\textwidth]{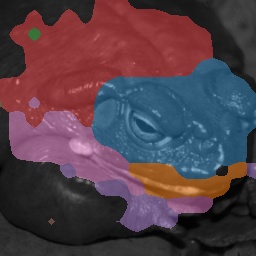}  &
 \includegraphics[width=0.1\textwidth]{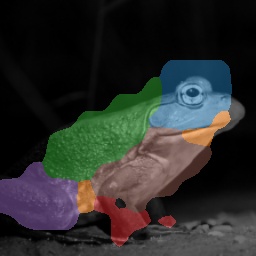} \\
 \rotatebox[origin=tl]{90}{{\parbox{0.95cm}{\centering K\\Medium}}} &
 \includegraphics[width=0.1\textwidth]{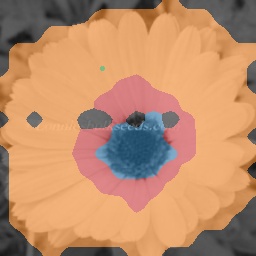} &
 \includegraphics[width=0.1\textwidth]{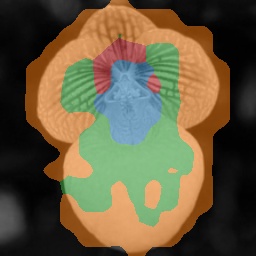}  &
 \includegraphics[width=0.1\textwidth]{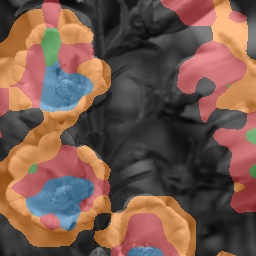}  &
 \includegraphics[width=0.1\textwidth]{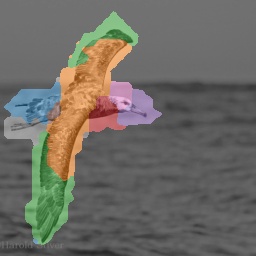}  &
 \includegraphics[width=0.1\textwidth]{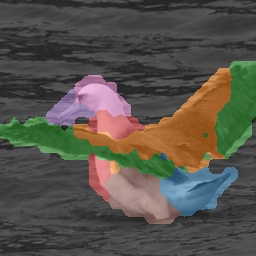}  &
 \includegraphics[width=0.1\textwidth]{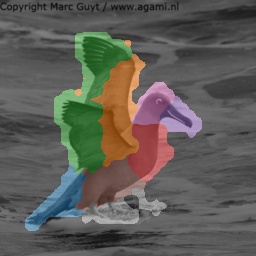}  &
 \includegraphics[width=0.1\textwidth]{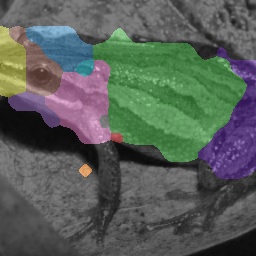}  &
 \includegraphics[width=0.1\textwidth]{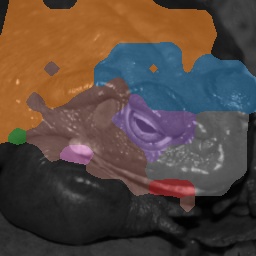}  &
 \includegraphics[width=0.1\textwidth]{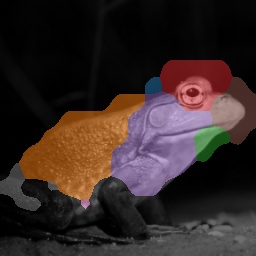} \\
\rotatebox[origin=tl]{90}{{\parbox{0.95cm}{\centering K\\High}}} &
\includegraphics[width=0.1\textwidth]{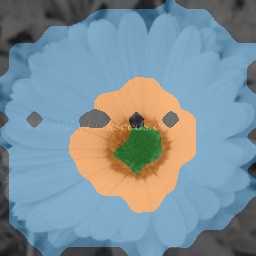} &
 \includegraphics[width=0.1\textwidth]{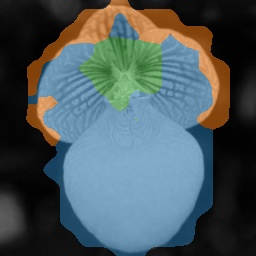}  &
 \includegraphics[width=0.1\textwidth]{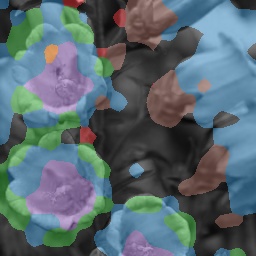}  &
 \includegraphics[width=0.1\textwidth]{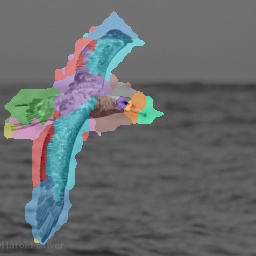}  &
 \includegraphics[width=0.1\textwidth]{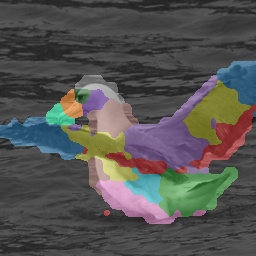}  &
 \includegraphics[width=0.1\textwidth]{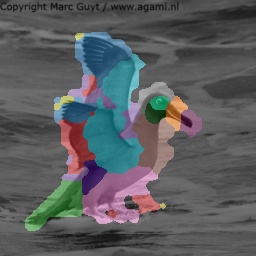}  &
 \includegraphics[width=0.1\textwidth]{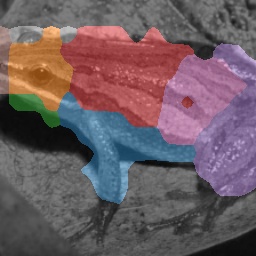}  &
 \includegraphics[width=0.1\textwidth]{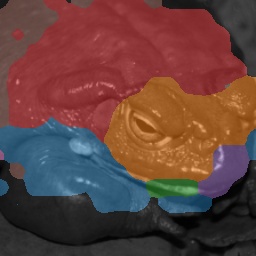}  &
 \includegraphics[width=0.1\textwidth]{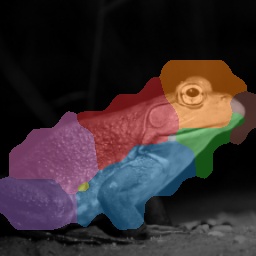} \\
 \end{tabular}
\end{ssmall}  
\caption{Effect of $K$ on PDiscoFormer. The left three images correspond to Flowers (from top to bottom, $K=2, 4, 8$). The three middle ones are CUB, with $K$ being 4, 8 and 16. The three on the right belong to PartImageNet, with $K=8, 25, 50$.}
\label{fig:k-variance}
\end{figure}

\noindent \textbf{Qualitative Analysis.}
\label{subsec:qual_analysis}
A qualitative analysis, as illustrated in~\cref{fig:qual-cub}, \cref{fig:qual-pimagenet}, \cref{fig:qual-flowers} and \cref{fig:k-variance} corroborates our quantitative findings, showcasing our model's ability to accurately discover irregularly shaped body parts such as bird wings (CUB), flower petals (Flowers) and the body of a snake (PartImageNet). 
At the same time, PDiscoNet~\cite{van2023pdisconet} fails to discover these parts altogether, even when retrained on a ViT-B backbone.
This indicates the usefulness of the flexible geometric priors enabled by our total variation loss. 
Compared to PDiscoNet~\cite{van2023pdisconet}, not only do we achieve a much tighter coverage of the object of interest, with our PDiscoFormer more closely following the object boundaries.
This is particularly the case in parts that are relatively large, such as open bird wings and flower corollas.
In addition to that, we can observe a substantial improvement in the consistency and semantic interpretability of parts. For instance, the parts discovered by our model for the four quadruped images displayed in~\cref{fig:qual-pimagenet} can immediately be interpreted (the green part corresponds to the snout while the brown one to the ears), while those from PDiscoNet are not clear.
In \cref{fig:k-variance} we vary the number of parts, $K$. On Flowers, using two parts leads to a clear assignment of one part to the flower calyx and one to the corolla, while more parts tends to exhibit oversegmentation.
CUB, on the other hand, is able to absorb a larger number of parts. With $K=4$, beak and legs are covered by the same part, while a fine-grained and consistent distinction of the different wing parts is achieved with $K=16$.
Interestingly, in the case of PartImageNet, the effective number of parts being used by a specific super-class remains almost constant when increasing $K$.


\begin{table}[t]
\centering
\caption{Results of the ablation study in PartImageNet and CUB.}
\label{tab:exp-ablation}
\begin{tabular}{c|ccc|cccc}
 &
  \multicolumn{3}{c|}{\begin{tabular}[c]{@{}c@{}}PartImageNet OOD \\ (K=25)\end{tabular}} &
  \multicolumn{4}{c}{\begin{tabular}[c]{@{}c@{}}CUB\\  (K=16)\end{tabular}} \\ \hline
 &
  NMI~$\uparrow$ &
  ARI~$\uparrow$ &
  \begin{tabular}[c]{@{}c@{}}Top-1\\ Acc.~$\uparrow$\end{tabular} &
  Kp~$\downarrow$ &
  NMI~$\uparrow$ &
  ARI~$\uparrow$ &
  \begin{tabular}[c]{@{}c@{}}Top-1\\ Acc.~$\uparrow$\end{tabular} \\ \hline
Full Model &
  \textbf{44.71} &
  \textbf{59.27} &
  90.77 &
  5.74 &
  \textbf{73.38} &
  \textbf{55.83} &
  88.72 \\ \hline
No $ \mathcal{L}_{\text{p}_{0}}$  & 41.05 & 49.08 & 89.81          & 5.85          & 68.66 & 43.09 & 83.64          \\
No $\mathcal{L}_{ent}$    & 39.16 & 54.46 & \textbf{91.19} & \textbf{5.57} & 66.88 & 45.68 & 88.21          \\
No $\mathcal{L}_{eq}$     & 43.38 & 54.19 & 90.47          & 9.05          & 56.85 & 30.63 & 83.71          \\
No $\mathcal{L}_{\perp}$      & 42.44 & 58.86 & 91.01          & 5.79          & 70.55 & 50.90 & 88.63          \\
No $ \mathcal{L}_{\text{p}_{1}}$  & 33.98 & 58.31 & 91.07          & 7.14          & 61.04 & 34.28 & \textbf{89.04} \\
No $\mathcal{L}_{tv}$     & 33.73 & 25.35 & 90.47          & 6.01          & 70.78 & 51.83 & 80.22          \\
No Per-Part LayerNorm  & 31.14 & 40.73 & 89.69          & 5.74          & 70.70 & 50.62 & 69.73          \\
No Part Dropout & 33.08 & 43.98 & 89.99          & 6.08          & 70.28 & 41.69 & 87.14          \\
No Gumbel       & 32.98 & 54.68 & 90.71          & 6.44          & 68.92 & 43.61 & 86.43         
\end{tabular}
\end{table}


\begin{figure}[ht]
\centering
\setlength\tabcolsep{1.5pt} 
\begin{ssmall}
 \begin{tabular}{cccccccccc}
  \rotatebox[origin=tl]{90}{{\parbox{0.95cm}{\centering Image}}} &
 \includegraphics[width=0.1\textwidth]{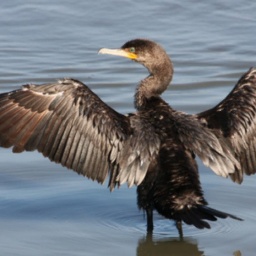} &
 \includegraphics[width=0.1\textwidth]{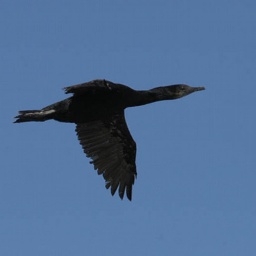}  &
 \includegraphics[width=0.1\textwidth]{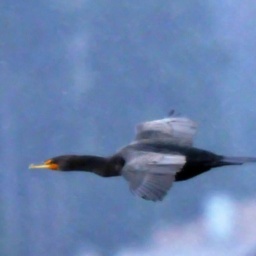}  &
 \includegraphics[width=0.1\textwidth]{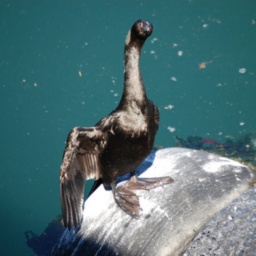}  &
 \includegraphics[width=0.1\textwidth]{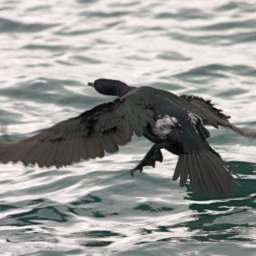}  &
 \includegraphics[width=0.1\textwidth]{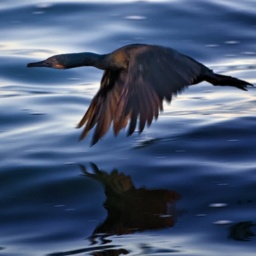}  &
 \includegraphics[width=0.1\textwidth]{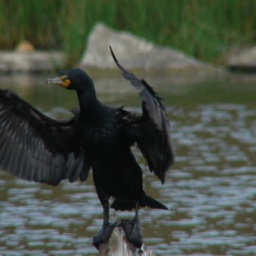}  &
 \includegraphics[width=0.1\textwidth]{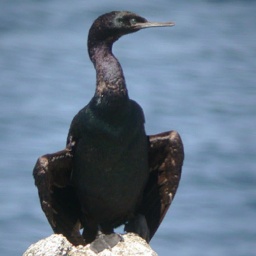}  &
 \includegraphics[width=0.1\textwidth]{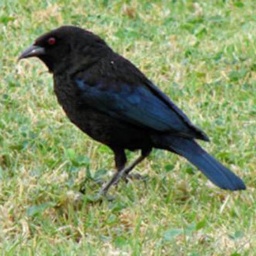}  \\
\rotatebox[origin=tl]{90}{{\parbox{0.95cm}{\centering Ours}}} &
     \includegraphics[width=0.1\textwidth]{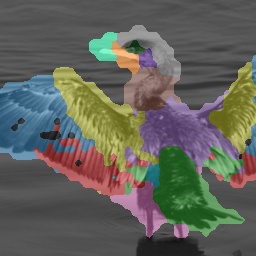} &
 \includegraphics[width=0.1\textwidth]{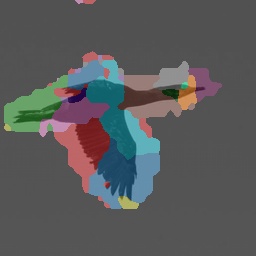}  &
 \includegraphics[width=0.1\textwidth]{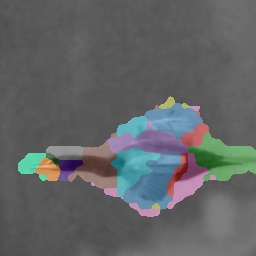}  &
 \includegraphics[width=0.1\textwidth]{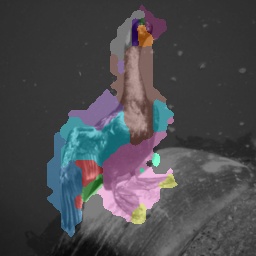}  &
 \includegraphics[width=0.1\textwidth]{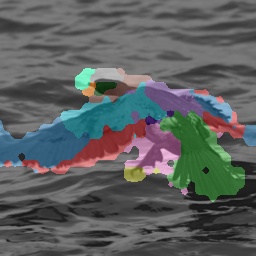}  &
 \includegraphics[width=0.1\textwidth]{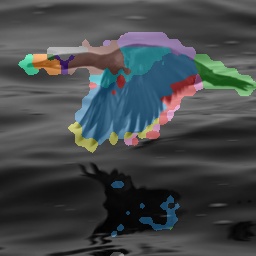}  &
 \includegraphics[width=0.1\textwidth]{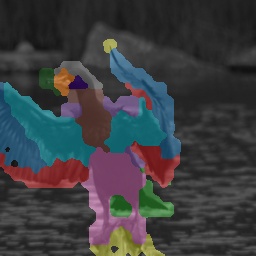}  &
 \includegraphics[width=0.1\textwidth]{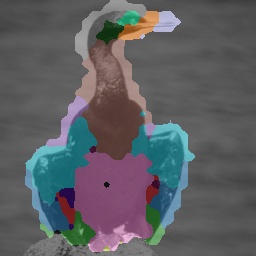}  &
 \includegraphics[width=0.1\textwidth]{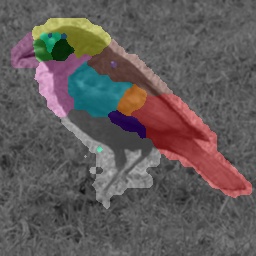} \\
\rotatebox[origin=tl]{90}{{\parbox{0.95cm}{\centering -$\mathcal{L}_{\text{p}_{0}}$}}} &
   \includegraphics[width=0.1\textwidth]{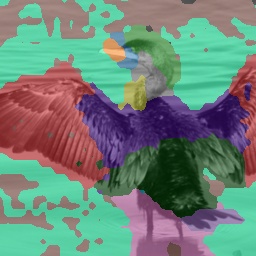} &
 \includegraphics[width=0.1\textwidth]{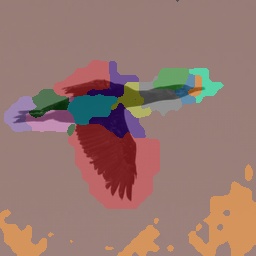}  &
 \includegraphics[width=0.1\textwidth]{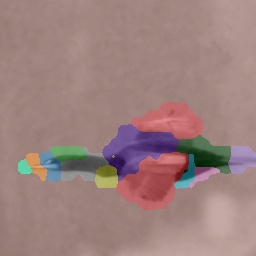}  &
 \includegraphics[width=0.1\textwidth]{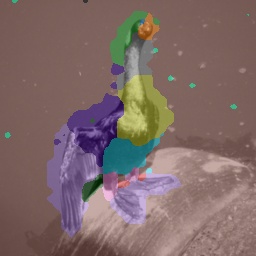}  &
 \includegraphics[width=0.1\textwidth]{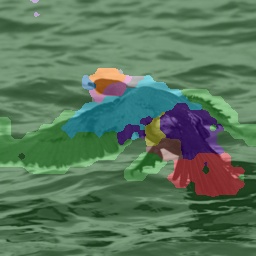}  &
 \includegraphics[width=0.1\textwidth]{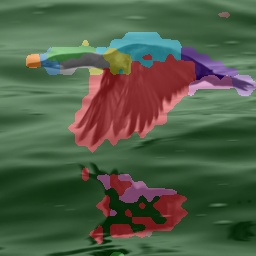}  &
 \includegraphics[width=0.1\textwidth]{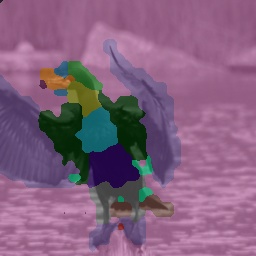}  &
 \includegraphics[width=0.1\textwidth]{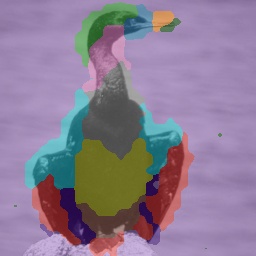}  &
 \includegraphics[width=0.1\textwidth]{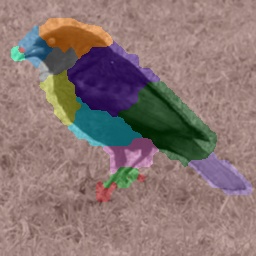} \\
 \rotatebox[origin=tl]{90}{{\parbox{0.95cm}{\centering -$\mathcal{L}_{\text{eq}}$}}} &
 \includegraphics[width=0.1\textwidth]{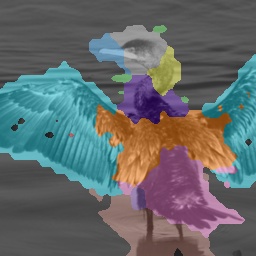} &
 \includegraphics[width=0.1\textwidth]{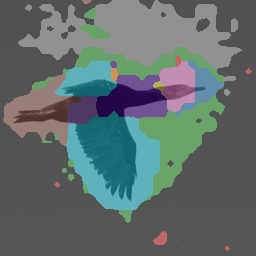}  &
 \includegraphics[width=0.1\textwidth]{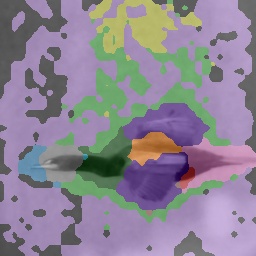}  &
 \includegraphics[width=0.1\textwidth]{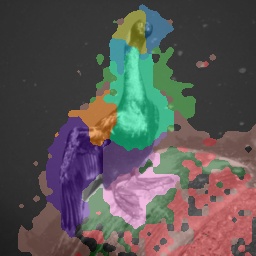}  &
 \includegraphics[width=0.1\textwidth]{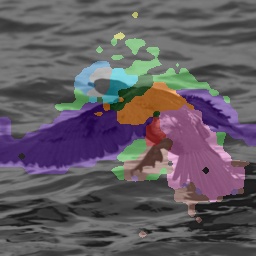}  &
 \includegraphics[width=0.1\textwidth]{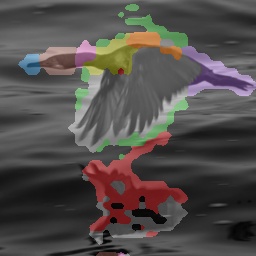}  &
 \includegraphics[width=0.1\textwidth]{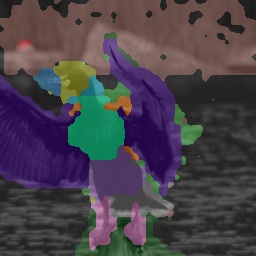}  &
 \includegraphics[width=0.1\textwidth]{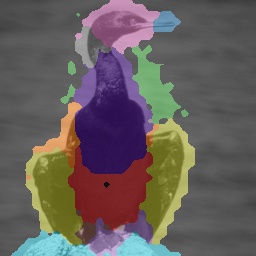}  &
 \includegraphics[width=0.1\textwidth]{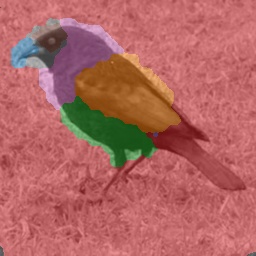}  \\
\rotatebox{90}{{\parbox{1.05cm}{\centering -Gumbel}}} &
 \includegraphics[width=0.1\textwidth]{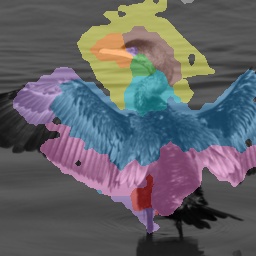} &
 \includegraphics[width=0.1\textwidth]{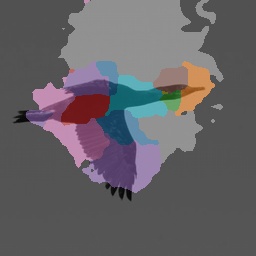}  &
 \includegraphics[width=0.1\textwidth]{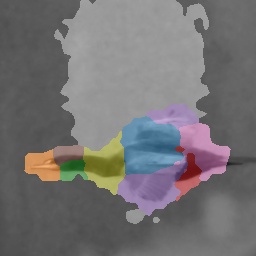}  &
 \includegraphics[width=0.1\textwidth]{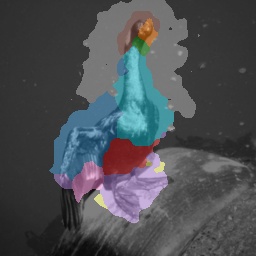}  &
 \includegraphics[width=0.1\textwidth]{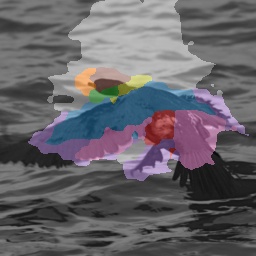}  &
 \includegraphics[width=0.1\textwidth]{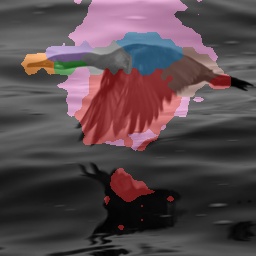}  &
 \includegraphics[width=0.1\textwidth]{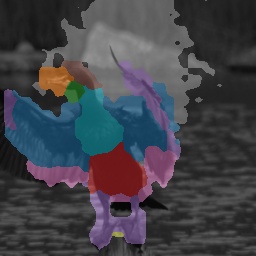}  &
 \includegraphics[width=0.1\textwidth]{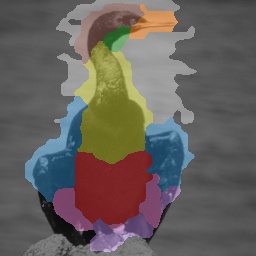}  &
 \includegraphics[width=0.1\textwidth]{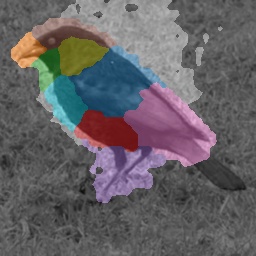} \\
\rotatebox{90}{{\parbox{1.05cm}{\centering -Perpart\\LN}}} &
\includegraphics[width=0.1\textwidth]{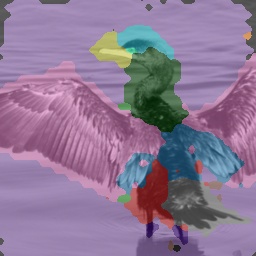} &
 \includegraphics[width=0.1\textwidth]{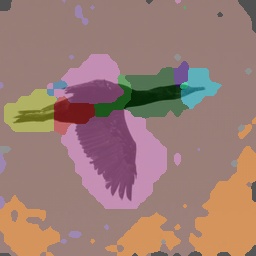}  &
 \includegraphics[width=0.1\textwidth]{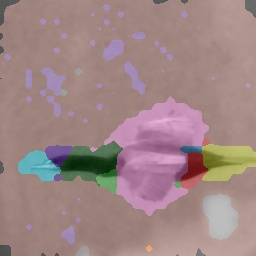}  &
 \includegraphics[width=0.1\textwidth]{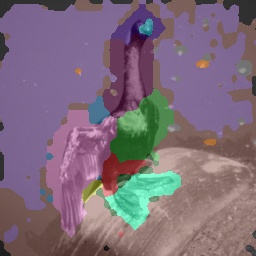}  &
 \includegraphics[width=0.1\textwidth]{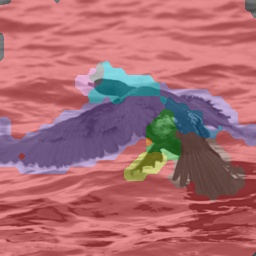}  &
 \includegraphics[width=0.1\textwidth]{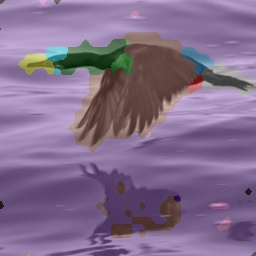}  &
 \includegraphics[width=0.1\textwidth]{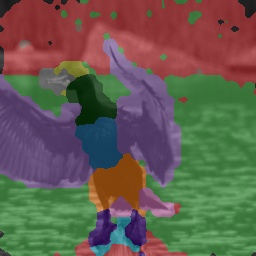}  &
 \includegraphics[width=0.1\textwidth]{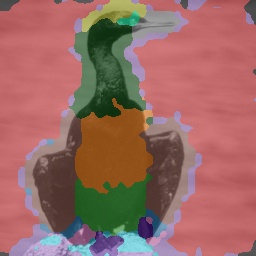}  &
 \includegraphics[width=0.1\textwidth]{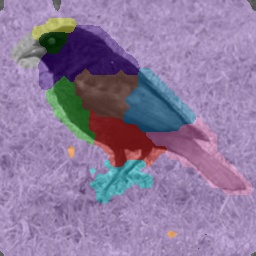} \\
 \rotatebox{90}{{\parbox{1.05cm}{\centering -Part\\Dropout}}} &
 \includegraphics[width=0.1\textwidth]{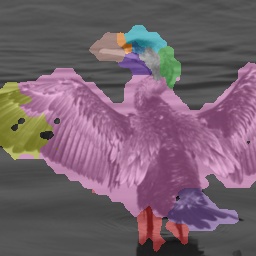} &
 \includegraphics[width=0.1\textwidth]{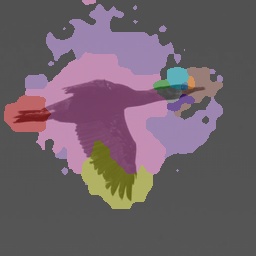}  &
 \includegraphics[width=0.1\textwidth]{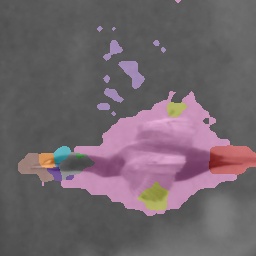}  &
 \includegraphics[width=0.1\textwidth]{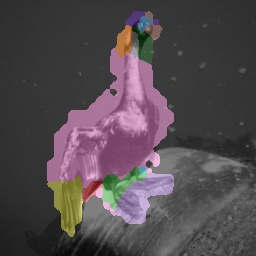}  &
 \includegraphics[width=0.1\textwidth]{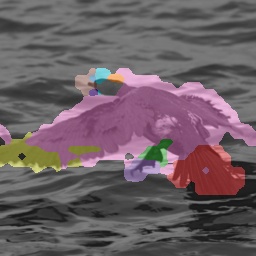}  &
 \includegraphics[width=0.1\textwidth]{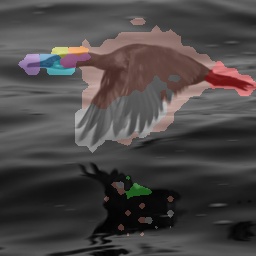}  &
 \includegraphics[width=0.1\textwidth]{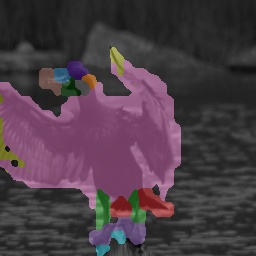}  &
 \includegraphics[width=0.1\textwidth]{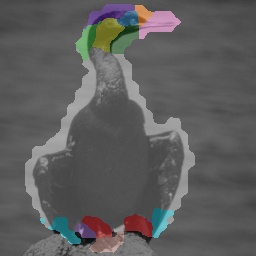}  &
 \includegraphics[width=0.1\textwidth]{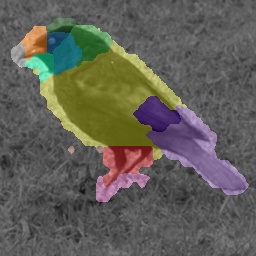}  \\
\rotatebox[origin=tl]{90}{{\parbox{0.95cm}{\centering -$\mathcal{L}_{tv}$}}} &
 \includegraphics[width=0.1\textwidth]{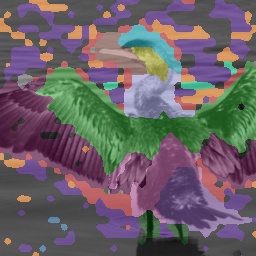} &
 \includegraphics[width=0.1\textwidth]{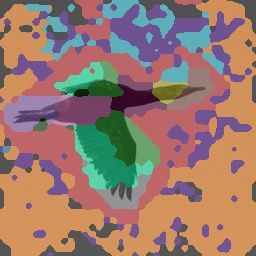}  &
 \includegraphics[width=0.1\textwidth]{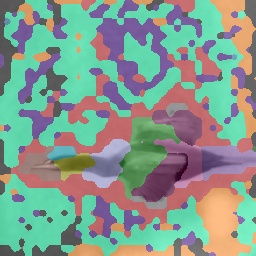}  &
 \includegraphics[width=0.1\textwidth]{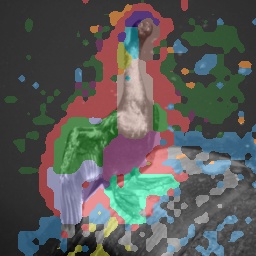}  &
 \includegraphics[width=0.1\textwidth]{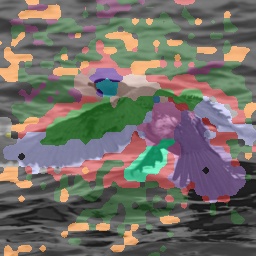}  &
 \includegraphics[width=0.1\textwidth]{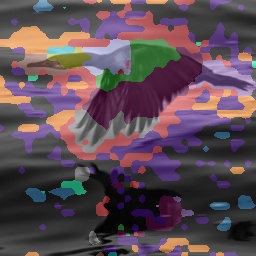}  &
 \includegraphics[width=0.1\textwidth]{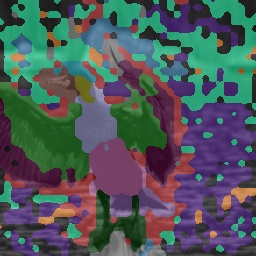}  &
 \includegraphics[width=0.1\textwidth]{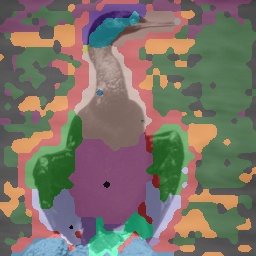}  &
 \includegraphics[width=0.1\textwidth]{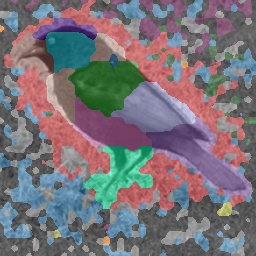} \\
 \end{tabular}    
\end{ssmall}
\caption{Qualitative results from the ablation study.}
\label{fig:qual-ablation}
\end{figure}


\subsection{Ablation Study}
\label{subsec:exp_ablation}
We conducted an ablation study by systematically removing one loss function or architectural element at a time and by replacing the Gumbel-Softmax with a regular Softmax. This was performed on the PartImageNet OOD dataset with $K=25$ parts and the CUB dataset with $K=16$ parts. The quantitative results of the ablation study are shown in \cref{tab:exp-ablation} and some qualitative results can be seen in \cref{fig:qual-ablation}. From the results, we can observe the following: \textbf{Total Variation ($\mathcal{L}_{tv}$)} prevents spurious part activations in the background, possibly reducing information leakage between foreground parts and the background. It has the largest impact on PartImageNet in terms of ARI. The \textbf{absence of the Background Loss ($\mathcal{L}_{\text{p}_{0}}$)} results in the background part not being activated, causing the model to discover the background as a foreground part. \textbf{Presence Loss ($\mathcal{L}_{\text{p}}$) and Part Dropout} promote the utilization of all parts, preventing domination by a few parts which can lead to multiple unused parts~(\cref{fig:qual-ablation}). \textbf{Removing the Per-Part Layer Norm} leads to inconsistent part discovery~(\cref{fig:qual-ablation}) indicating that it contributes to the creation of discriminative part-specific features driven by the classification loss. \textbf{Equivariance Loss ($\mathcal{L}_{eq}$)} enhances robustness to affine geometric transformations, supporting stable part discovery across diverse inputs~(\cref{fig:qual-ablation}). \textbf{Gumbel-Softmax and Entropy Loss ($\mathcal{L}_{\text{ent}}$)} ensure unique part assignments for each patch token, further reducing information leakage, as shown in~\cref{sec-app:entropy}. As shown in \cref{tab:exp-ablation}, the complete model, incorporating all loss functions and architectural choices, generally performs best for both part discovery and classification, demonstrating the efficacy of our approach.

\section{Limitations and Future Work}
\label{sec:limitations}

Our study focused on datasets with part annotations or FG-BG masks available for the test set. 
This allowed for quantitative comparison with state-of-the-art methods. However, exploring the training of our model on larger and more diverse datasets could provide valuable insights and further validate its performance in real-world scenarios. For example, larger datasets may introduce more variability in object appearances and poses, challenging the model to generalize better. Methods to enable the evaluation of part discovery in such settings would allow to take steps in this direction.
The main hyper-parameter \(K\), the number of parts to discover, has a large impact on the granularity of the discovered parts (\cref{fig:k-variance}). Exploring ways of automatically estimating this value while preserving interpretability in the discovered semantic parts would be a valuable contribution. This could enhance the adaptability of the model to various datasets and streamline the training process by removing the need for manual selection of \(K\).
Additionally, while our approach can detect parts of various shapes due to the total variation prior, it cannot ascertain if they belong to the same object. Integrating our method with unsupervised object discovery research \cite{simeoni2021localizing,henaff2022object} could enable part discovery within each detected object, improving capabilities in complex scenes. These methods aim to identify and localize individual objects without explicit supervision, complementing our approach for a more comprehensive scene understanding. These considerations suggest promising directions for extending and refining our method.

\section{Conclusion}
\label{sec:conclusion}
In this paper, we presented PDiscoFormer, a new approach for unsupervised part discovery. Unlike previous methods that imposed strict geometric priors on discovered object parts, our model, based on self-supervised vision transformers, demonstrated the capability to relax these constraints. By leveraging a total variation (TV) prior, our model achieved substantial improvements in both part discovery and downstream classification accuracy when compared to the state-of-the-art across multiple datasets, specifically CUB, PartImageNet, and Oxford Flowers.
Our findings highlight the importance of rethinking geometric priors in unsupervised part discovery tasks, especially in scenarios where objects exhibit complex shapes or occur multiple times within the same image. By refraining from assuming specific shape or size priors, our model demonstrated enhanced generalization capabilities, making it applicable to a broader range of fine-grained image classification tasks.
Overall, our study shows that the strong inductive biases learned by vision transformers during the self-supervised pre-training stage enable the use of flexible shape priors such as total variation, leading to robust performance on part discovery and the downstream classification task on all of the tested datasets.

\noindent \textbf{Acknowledgment.} This work was supported in part by ANR project OBTEA ANR-22-CPJ1-0054-01 and, was granted access to the HPC resources of IDRIS under the allocation 2023-AD011014325 made by GENCI.
%
%
\bibliographystyle{splncs04}
\bibliography{egbib}
\clearpage
\appendix
\begin{table}[t]
    \centering
    \caption{Training details for PDiscoFormer on different datasets.}
    \label{tab:training_details}
    \begin{tabular}{lcc}
        \hline
        Dataset & Batch Size per GPU & Training Time \\
        \hline
        CUB & 8 & 4 hours \\
        PartImageNet OOD & 32 & 1 hour 10 minutes \\
        Flowers & 32 & 14 minutes \\
        \hline
    \end{tabular}
\end{table}
\begin{table}[t]
    \centering
    \caption{Inference speed comparison on the CUB dataset with $K=8$.}
    \label{tab:inference_speed}
    \begin{tabular}{lc}
        \hline
        Model & Inference Speed (images/second) \\
        \hline
        Huang~\cite{Huang2020InterpretableGrouping} & 139.31 \\
        PDiscoNet~\cite{van2023pdisconet} & 248.07 \\
        PDiscoFormer & \textbf{326.18} \\
        \hline
    \end{tabular}
\end{table}
\section{Training Settings}
\label{sec-app:train_settings}
We trained all our models using the Adam optimizer \cite{Kingma2015Adam:Optimization}. The class token, position embedding, and register token of the ViT were kept unfrozen, while all other ViT layers were frozen during training. We used a starting learning rate of $10^{-6}$ for the ViT backbone fine-tuned tokens, $10^{-3}$ for the linear projection layer to form the part prototypes, and $10^{-2}$ for the modulation and the final linear layer used for classification. We used a variable batch size, with a minimum of 16, by adjusting the learning rate using the square root scaling rule~\cite{krizhevsky2014one}. Training lasted for a total of 28 epochs, and we employed a step learning rate schedule, reducing the learning rate by a factor of 0.5 every 4 epochs (as in~\cite{van2023pdisconet}). Additionally, to regularize our training process, we applied gradient norm clipping \cite{Pascanu2013OnNetworks} with a constant value of 2 for all experiments.
In all our experiments, the loss weight of the background loss $\mathcal{L}_{\text{p}_{0}}$ was set to 2, while all other loss weights were set to a value of 1. We used a constant part dropout value of 0.3.

\section{Compute Requirements and Model Performance}
\label{sec-app:compute_req}
\noindent \textbf{Compute Requirements (Training).}
We trained our models on a machine equipped with 4 NVIDIA GeForce RTX 2080 Ti GPUs. The training duration varied depending on the dataset and batch size, as detailed in~\cref{tab:training_details}.

\noindent \textbf{Inference Speed.}
Taking the models trained on the CUB dataset~\cite{WelinderEtal2010} with $K=8$ as an example, the inference speed on a workstation with a single NVIDIA GeForce RTX 3090 GPU, averaged over 100 runs, is presented in~\cref{tab:inference_speed}.

\begin{table}[t]
\centering
\caption{Performance of our method and the state-of-the-art method from the literature~\cite{van2023pdisconet} under different backbone configurations.}
\label{tab:sup-results}
\begin{adjustbox}{width=\linewidth}
\begin{tabular}{c|ccccc|cccc|ccc}
& \multicolumn{5}{c|}{CUB (\%)}                                                                                                    & \multicolumn{4}{c|}{PartImageNet OOD (\%)}                                                                       & \multicolumn{3}{c}{Flowers (\%)}                                                                                                    \\ \cline{2-13} 
 Method  & \multicolumn{1}{c|}{K}  & Kp~$\downarrow$            & NMI~$\uparrow$            & ARI~$\uparrow$            & \begin{tabular}[c]{@{}c@{}}Top-1\\ Acc.~$\uparrow$\end{tabular} & \multicolumn{1}{c|}{K}  & NMI~$\uparrow$            & ARI~$\uparrow$            & \begin{tabular}[c]{@{}c@{}}Top-1\\ Acc.~$\uparrow$\end{tabular} & \multicolumn{1}{c|}{K} & \begin{tabular}[c]{@{}c@{}}Fg. \\ mIoU~$\uparrow$\end{tabular} & \begin{tabular}[c]{@{}c@{}}Top-1\\ Acc.~$\uparrow$\end{tabular} \\ \hline
\multirow{3}{*}{\begin{tabular}[c]{@{}c@{}}PDiscoNet~\cite{van2023pdisconet}\\ + R101\end{tabular}}   & \multicolumn{1}{c|}{4}  & 9.12          & 37.82          & 15.26          & 86.17                                                & \multicolumn{1}{c|}{8}  & 27.13          & 8.76           & 88.58                                                & \multicolumn{1}{c|}{2} & 19.04                                               & 77.51                                                \\
                                                                              & \multicolumn{1}{c|}{8}  & 8.52          & 50.08          & 26.96          & 86.72                                                & \multicolumn{1}{c|}{25} & 32.41          & 10.69          & 89.00                                                & \multicolumn{1}{c|}{4} & 34.76                                               & 83.05                                                \\
                                                                              & \multicolumn{1}{c|}{16} & 7.67          & 56.87          & 38.05          & 87.49                                                & \multicolumn{1}{c|}{50} & 41.49          & 14.17          & 86.06                                                & \multicolumn{1}{c|}{8} & 49.10                                               & 81.04                                                \\ \hline
\multirow{3}{*}{\begin{tabular}[c]{@{}c@{}}PDiscoFormer\\ + R101\end{tabular}}        & \multicolumn{1}{c|}{4}  & 11.07         & 34.32          & 16.94          & 82.59                                                & \multicolumn{1}{c|}{8}  & 11.44          & 29.64          & 87.33                                                & \multicolumn{1}{c|}{2} & 0.89                                                & 8.41                                                 \\
                                                                              & \multicolumn{1}{c|}{8}  & 8.27          & 44.59          & 25.63          & 84.25                                                & \multicolumn{1}{c|}{25} & 13.86          & 27.55          & 88.42                                                & \multicolumn{1}{c|}{4} & 0.08                                                & 7.12                                                 \\
                                                                              & \multicolumn{1}{c|}{16} & 9.53          & 35.64          & 17.61          & 83.84                                                & \multicolumn{1}{c|}{50} & 7.91           & 19.64          & 88.78                                                & \multicolumn{1}{c|}{8} & 0.00                                                & 4.96                                                 \\ \hline \hline
\multirow{3}{*}{\begin{tabular}[c]{@{}c@{}}PDiscoNet\\ + ViT-B\end{tabular}}  & \multicolumn{1}{c|}{4}  & 7.70          & 52.59          & 26.66          & 88.61                                                & \multicolumn{1}{c|}{8}  & 19.28          & 34.72          & 90.95                                                & \multicolumn{1}{c|}{2} & 4.92                                                & 92.75                                                \\
                                                                              & \multicolumn{1}{c|}{8}  & 6.34          & 65.01          & 37.90          & 86.95                                                & \multicolumn{1}{c|}{25} & 28.23          & 50.35          & 90.29                                                & \multicolumn{1}{c|}{4} & 1.95                                                & 95.48                                                \\
                                                                              & \multicolumn{1}{c|}{16} & 5.95          & 68.63          & 43.41          & 84.04                                                & \multicolumn{1}{c|}{50} & 29.48          & 27.80          & 89.69                                                & \multicolumn{1}{c|}{8} & 13.18                                               & 97.40                                                \\ \hline
\multirow{3}{*}{\begin{tabular}[c]{@{}c@{}}PDiscoFormer + \\ frozen ViT-B\end{tabular}} & \multicolumn{1}{c|}{4}  & 8.19          & 52.88          & 23.22          & 88.87                                                & \multicolumn{1}{c|}{8}  & 28.84          & 55.66          & 90.35                                                & \multicolumn{1}{c|}{2} & 67.28                                               & 99.41                                                \\
                                                                              & \multicolumn{1}{c|}{8}  & 6.23          & 67.59          & 41.35          & 88.56                                                & \multicolumn{1}{c|}{25} & 43.36          & 62.82          & 90.47                                                & \multicolumn{1}{c|}{4} & 58.71                                               & 99.43                                                \\
                                                                              & \multicolumn{1}{c|}{16} & 6.44          & 69.54          & 49.99          & 85.10                                                & \multicolumn{1}{c|}{50} & 44.48          & 57.91          & 90.59                                                & \multicolumn{1}{c|}{8} & 72.27                                               & 99.53                                                \\ \hline
\multirow{3}{*}{\begin{tabular}[c]{@{}c@{}}PDiscoFormer +\\partially fine-tuned \\ViT-B\end{tabular}}       & \multicolumn{1}{c|}{4}  & 7.41          & 58.13          & 25.11          & 89.06                                                & \multicolumn{1}{c|}{8}  & 29.00          & 52.40          & 89.75                                                & \multicolumn{1}{c|}{2} & \textbf{73.62}                                      & 99.61                                                \\
                                                                              & \multicolumn{1}{c|}{8}  & 6.54          & 69.87          & 42.76          & \textbf{89.41}                                       & \multicolumn{1}{c|}{25} & 44.71          & 59.27          & 90.77                                                & \multicolumn{1}{c|}{4} & 73.32                                               & 99.54                                                \\
                                                                              & \multicolumn{1}{c|}{16} & \textbf{5.74} & \textbf{73.38} & \textbf{55.83} & 88.72                                                & \multicolumn{1}{c|}{50} & \textbf{46.29} & \textbf{62.21} & \textbf{91.01}                                       & \multicolumn{1}{c|}{8} & 69.59                                               & \textbf{99.64}                                      
\end{tabular}
\end{adjustbox}
\end{table}

\section{Effect of backbone}
\label{sec-app:backbone}
\begin{figure}[t]
\centering
\setlength\tabcolsep{1.5pt} 
\begin{ssmall}
\centering
 \begin{tabular}{cccccccccc}
  \rotatebox[origin=tl]{90}{{\parbox{1.05cm}{\centering Image}}} &
 \includegraphics[width=0.1\textwidth]{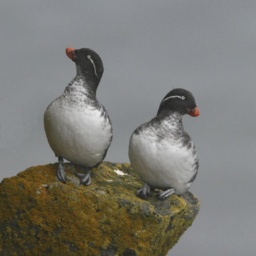} &
 \includegraphics[width=0.1\textwidth]{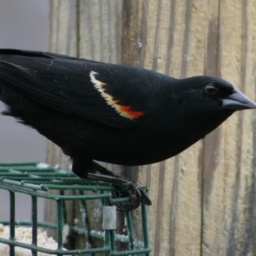}  &
 \includegraphics[width=0.1\textwidth]{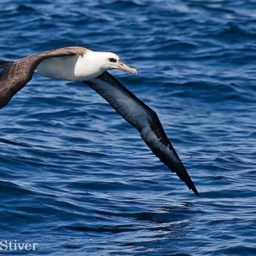}  &
 \includegraphics[width=0.1\textwidth]{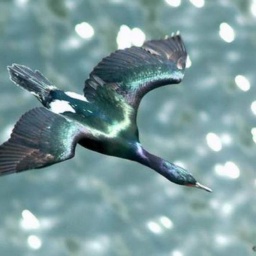}  &
 \includegraphics[width=0.1\textwidth]{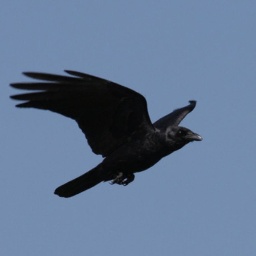}  &
 \includegraphics[width=0.1\textwidth]{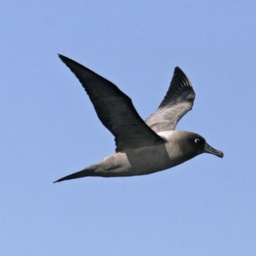}  &
 \includegraphics[width=0.1\textwidth]{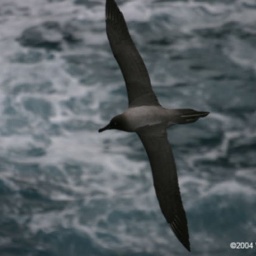}  &
 \includegraphics[width=0.1\textwidth]{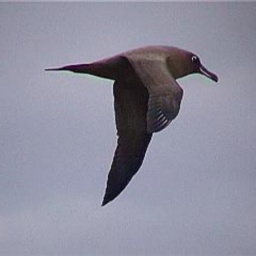}  &
 \includegraphics[width=0.1\textwidth]{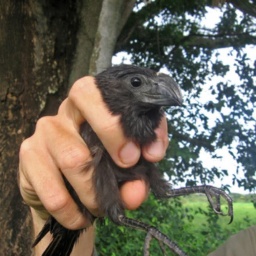}  \\
\rotatebox{90}{{\parbox{1.05cm}{\centering PDiscoNet\\+R101}}} &
     \includegraphics[width=0.1\textwidth]{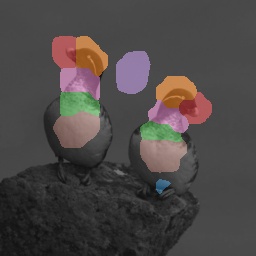} &
 \includegraphics[width=0.1\textwidth]{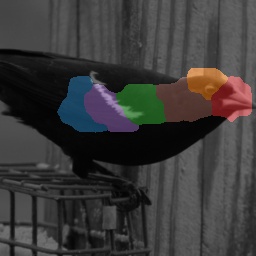}  &
 \includegraphics[width=0.1\textwidth]{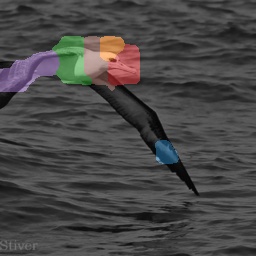}  &
 \includegraphics[width=0.1\textwidth]{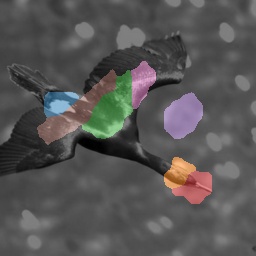}  &
 \includegraphics[width=0.1\textwidth]{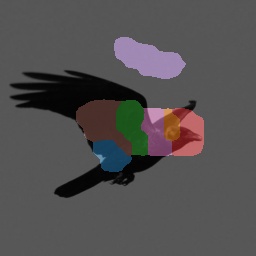}  &
 \includegraphics[width=0.1\textwidth]{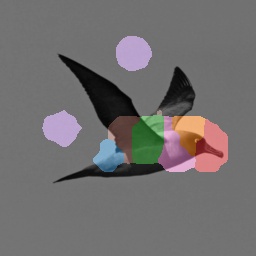}  &
 \includegraphics[width=0.1\textwidth]{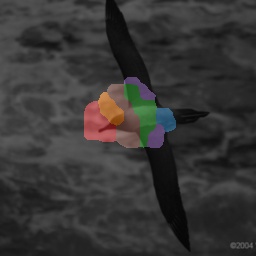}  &
 \includegraphics[width=0.1\textwidth]{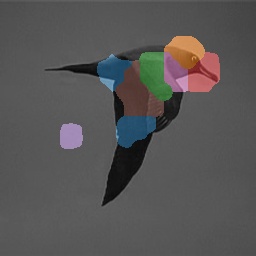}  &
 \includegraphics[width=0.1\textwidth]{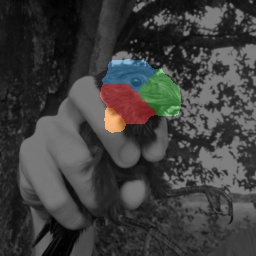} \\
 \rotatebox{90}{{\parbox{1.05cm}{\centering Ours \\+ R101}}} &
     \includegraphics[width=0.1\textwidth]{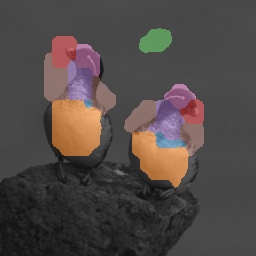} &
 \includegraphics[width=0.1\textwidth]{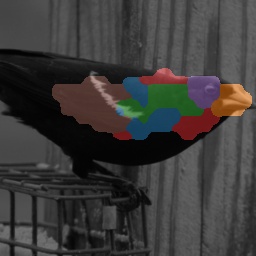}  &
 \includegraphics[width=0.1\textwidth]{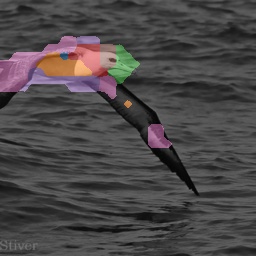}  &
 \includegraphics[width=0.1\textwidth]{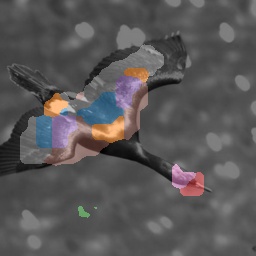}  &
 \includegraphics[width=0.1\textwidth]{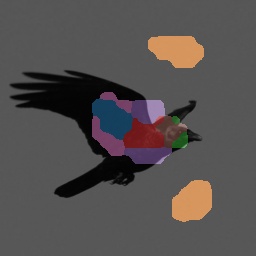}  &
 \includegraphics[width=0.1\textwidth]{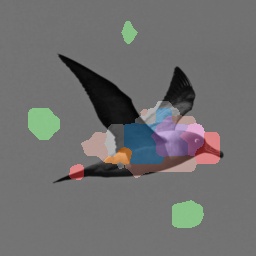}  &
 \includegraphics[width=0.1\textwidth]{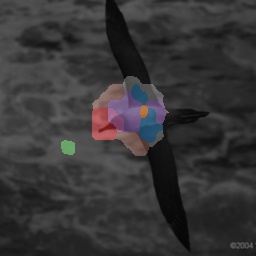}  &
 \includegraphics[width=0.1\textwidth]{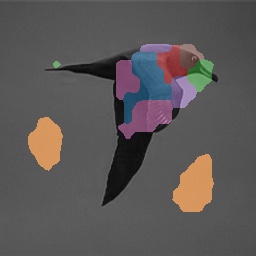}  &
 \includegraphics[width=0.1\textwidth]{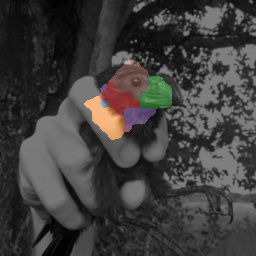} \\
 \rotatebox{90}{{\parbox{1.05cm}{\centering PDiscoNet + ViT-B}}} &
     \includegraphics[width=0.1\textwidth]{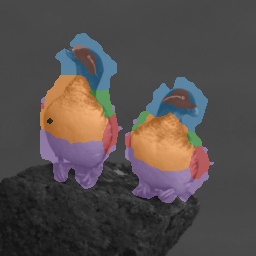} &
 \includegraphics[width=0.1\textwidth]{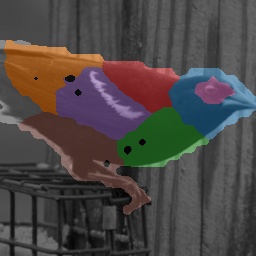}  &
 \includegraphics[width=0.1\textwidth]{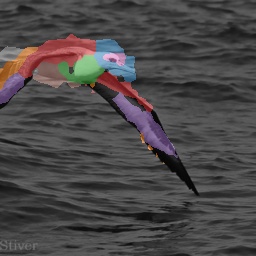}  &
 \includegraphics[width=0.1\textwidth]{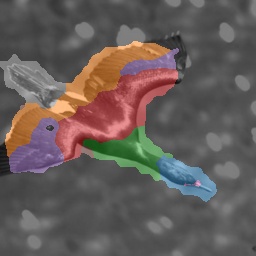}  &
 \includegraphics[width=0.1\textwidth]{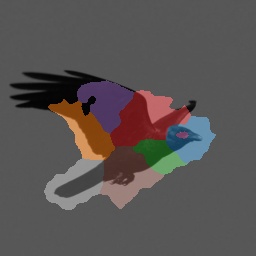}  &
 \includegraphics[width=0.1\textwidth]{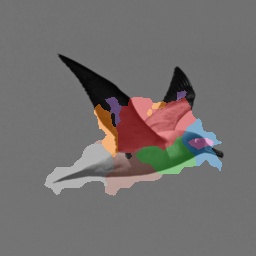}  &
 \includegraphics[width=0.1\textwidth]{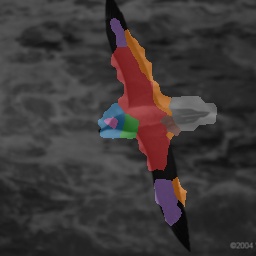}  &
 \includegraphics[width=0.1\textwidth]{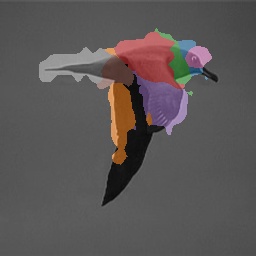}  &
 \includegraphics[width=0.1\textwidth]{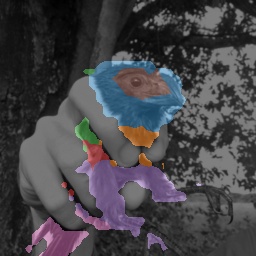} \\
 \rotatebox{90}{{\parbox{1.05cm}{\centering Ours \\+ ViT-B (frozen)}}} &
     \includegraphics[width=0.1\textwidth]{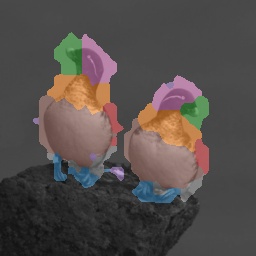} &
 \includegraphics[width=0.1\textwidth]{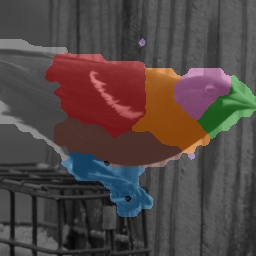}  &
 \includegraphics[width=0.1\textwidth]{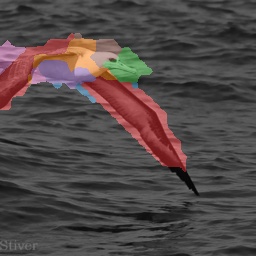}  &
 \includegraphics[width=0.1\textwidth]{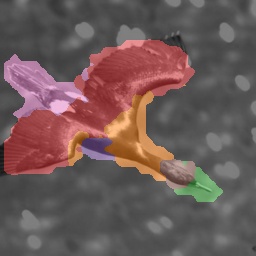}  &
 \includegraphics[width=0.1\textwidth]{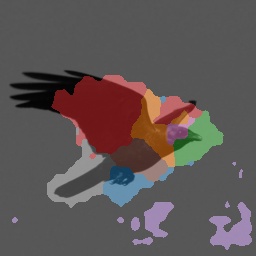}  &
 \includegraphics[width=0.1\textwidth]{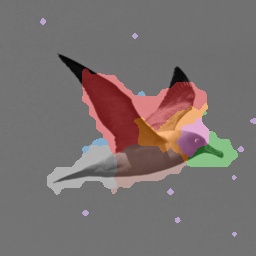}  &
 \includegraphics[width=0.1\textwidth]{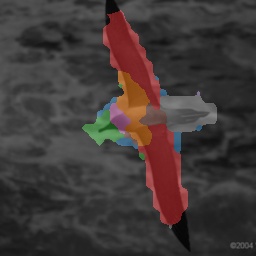}  &
 \includegraphics[width=0.1\textwidth]{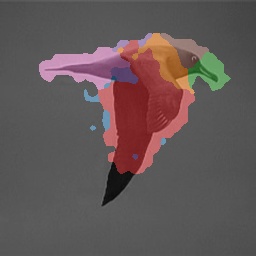}  &
 \includegraphics[width=0.1\textwidth]{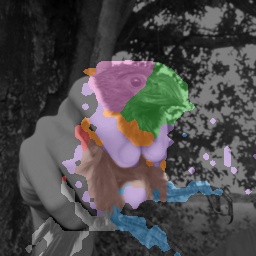} \\
\rotatebox[origin=tl]{90}{{\parbox{0.95cm}{\centering Ours}}} &
   \includegraphics[width=0.1\textwidth]{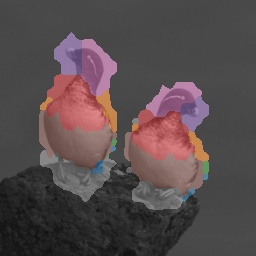} &
 \includegraphics[width=0.1\textwidth]{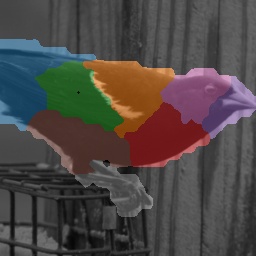}  &
 \includegraphics[width=0.1\textwidth]{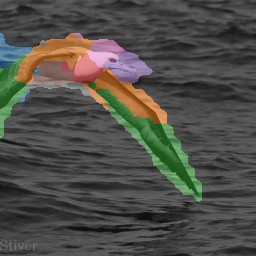}  &
 \includegraphics[width=0.1\textwidth]{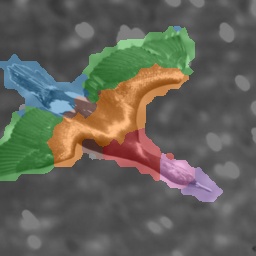}  &
 \includegraphics[width=0.1\textwidth]{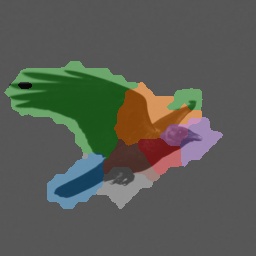}  &
 \includegraphics[width=0.1\textwidth]{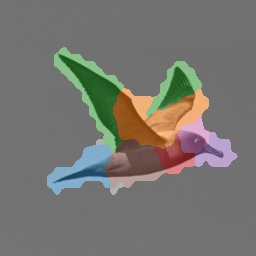}  &
 \includegraphics[width=0.1\textwidth]{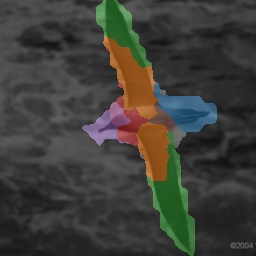}  &
 \includegraphics[width=0.1\textwidth]{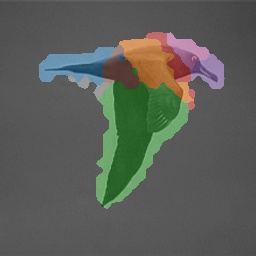}  &
 \includegraphics[width=0.1\textwidth]{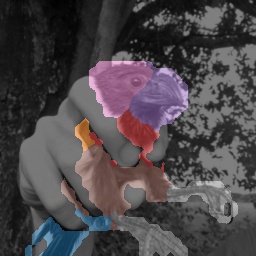} \\
\end{tabular}
\end{ssmall}
\caption{Qualitative results on CUB for $K=8$.}
\label{fig:qual-cub-supp}
\end{figure}
\begin{figure}[t]
\centering
\setlength\tabcolsep{1.5pt} 
\begin{ssmall}
\centering
 \begin{tabular}{cccccccccc}
  \rotatebox[origin=tl]{90}{{\parbox{1.05cm}{\centering Image}}} &
 \includegraphics[width=0.1\textwidth]{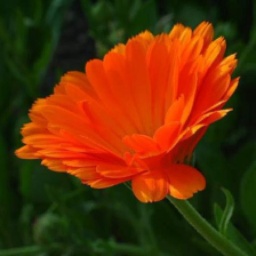} &
 \includegraphics[width=0.1\textwidth]{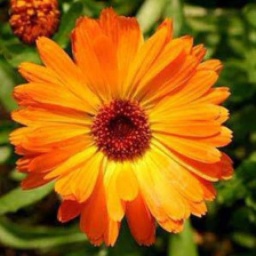}  &
 \includegraphics[width=0.1\textwidth]{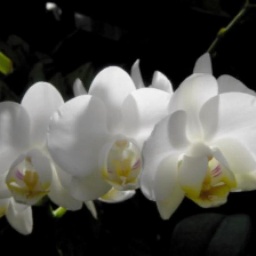}  &
 \includegraphics[width=0.1\textwidth]{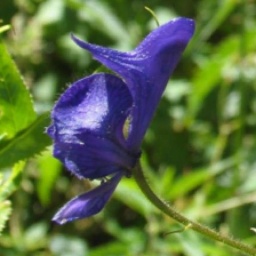}  &
 \includegraphics[width=0.1\textwidth]{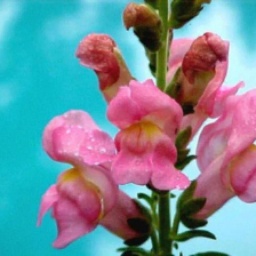}  &
 \includegraphics[width=0.1\textwidth]{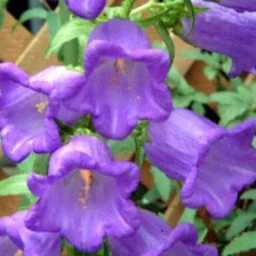}  &
 \includegraphics[width=0.1\textwidth]{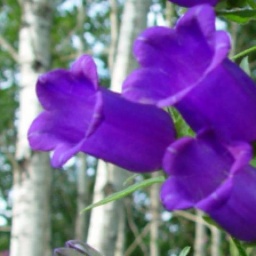}  &
 \includegraphics[width=0.1\textwidth]{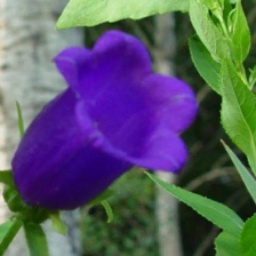}  &
 \includegraphics[width=0.1\textwidth]{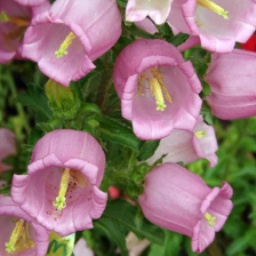}  \\
\rotatebox{90}{{\parbox{1.05cm}{\centering PDiscoNet\\+R101}}} &
     \includegraphics[width=0.1\textwidth]{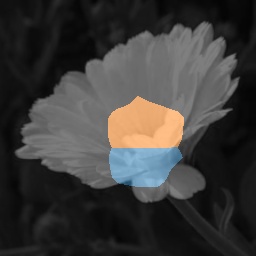} &
 \includegraphics[width=0.1\textwidth]{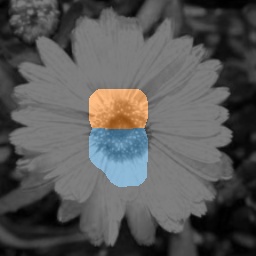}  &
 \includegraphics[width=0.1\textwidth]{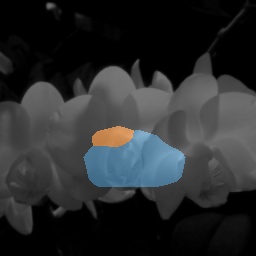}  &
 \includegraphics[width=0.1\textwidth]{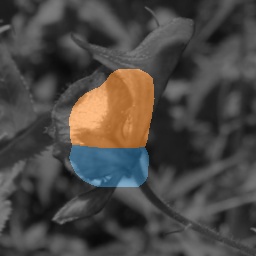}  &
 \includegraphics[width=0.1\textwidth]{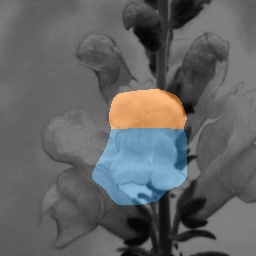}  &
 \includegraphics[width=0.1\textwidth]{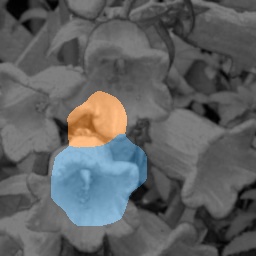}  &
 \includegraphics[width=0.1\textwidth]{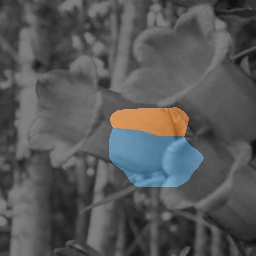}  &
 \includegraphics[width=0.1\textwidth]{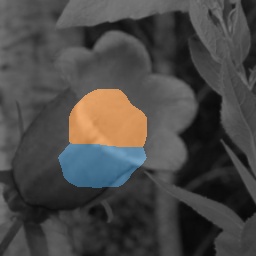}  &
 \includegraphics[width=0.1\textwidth]{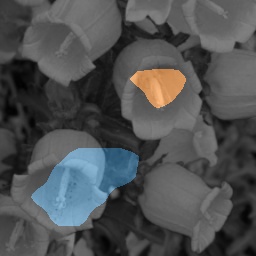} \\
 \rotatebox{90}{{\parbox{1.05cm}{\centering Ours \\+ R101}}} &
     \includegraphics[width=0.1\textwidth]{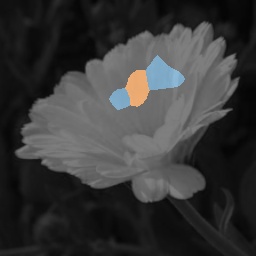} &
 \includegraphics[width=0.1\textwidth]{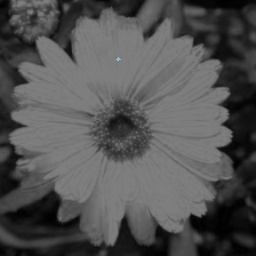}  &
 \includegraphics[width=0.1\textwidth]{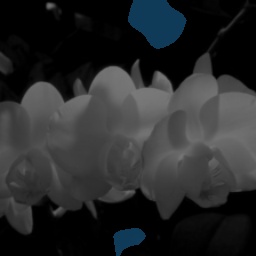}  &
 \includegraphics[width=0.1\textwidth]{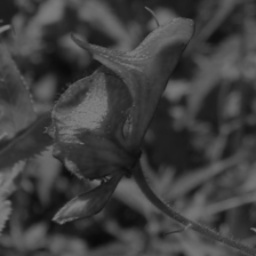}  &
 \includegraphics[width=0.1\textwidth]{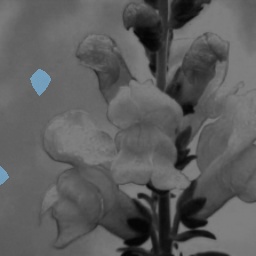}  &
 \includegraphics[width=0.1\textwidth]{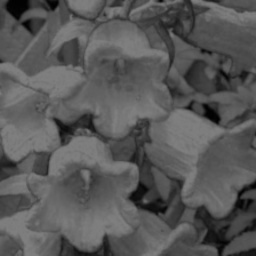}  &
 \includegraphics[width=0.1\textwidth]{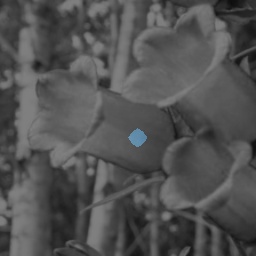}  &
 \includegraphics[width=0.1\textwidth]{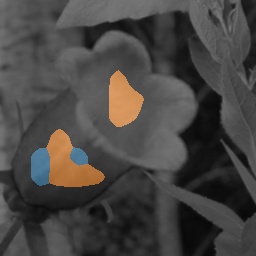}  &
 \includegraphics[width=0.1\textwidth]{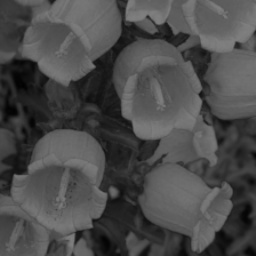} \\
 \rotatebox{90}{{\parbox{1.05cm}{\centering PDiscoNet + ViT-B}}} &
     \includegraphics[width=0.1\textwidth]{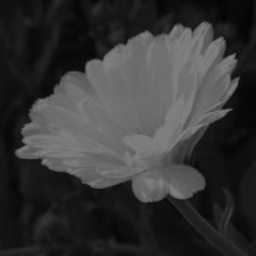} &
 \includegraphics[width=0.1\textwidth]{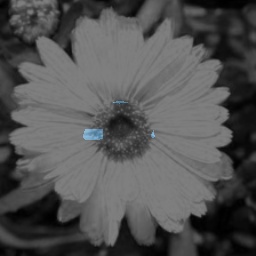}  &
 \includegraphics[width=0.1\textwidth]{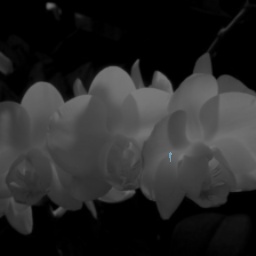}  &
 \includegraphics[width=0.1\textwidth]{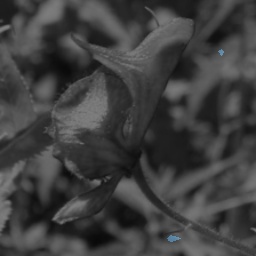}  &
 \includegraphics[width=0.1\textwidth]{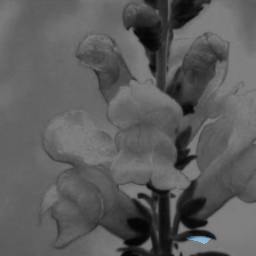}  &
 \includegraphics[width=0.1\textwidth]{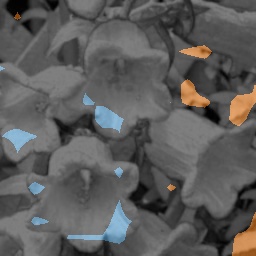}  &
 \includegraphics[width=0.1\textwidth]{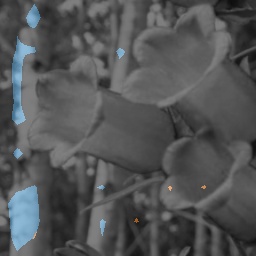}  &
 \includegraphics[width=0.1\textwidth]{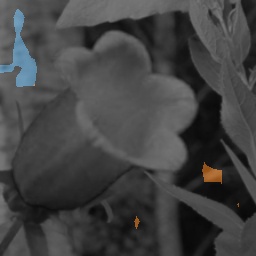}  &
 \includegraphics[width=0.1\textwidth]{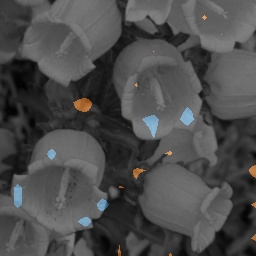} \\
 \rotatebox{90}{{\parbox{1.05cm}{\centering Ours \\+ ViT-B (frozen)}}} &
     \includegraphics[width=0.1\textwidth]{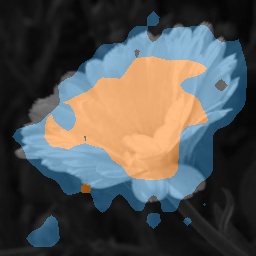} &
 \includegraphics[width=0.1\textwidth]{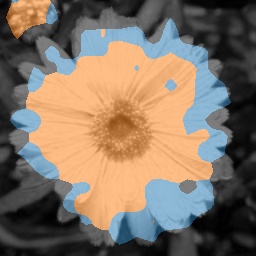}  &
 \includegraphics[width=0.1\textwidth]{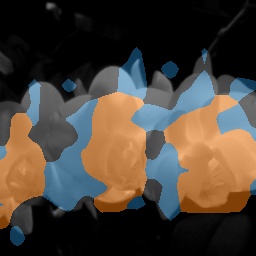}  &
 \includegraphics[width=0.1\textwidth]{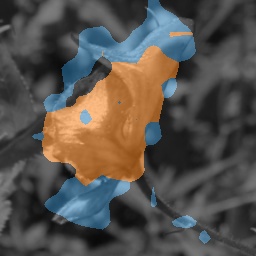}  &
 \includegraphics[width=0.1\textwidth]{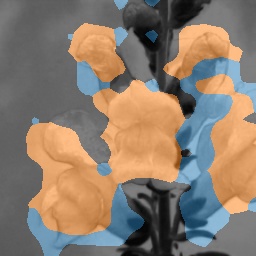}  &
 \includegraphics[width=0.1\textwidth]{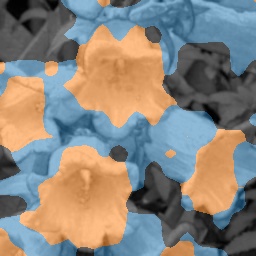}  &
 \includegraphics[width=0.1\textwidth]{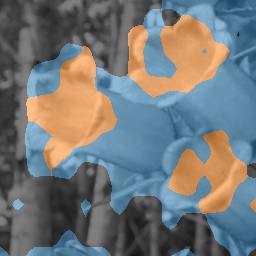}  &
 \includegraphics[width=0.1\textwidth]{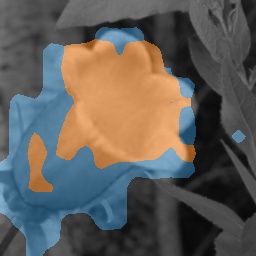}  &
 \includegraphics[width=0.1\textwidth]{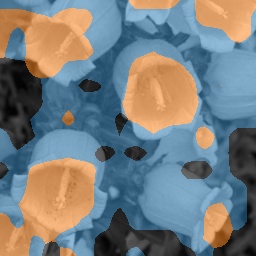} \\
\rotatebox[origin=tl]{90}{{\parbox{0.95cm}{\centering Ours}}} &
   \includegraphics[width=0.1\textwidth]{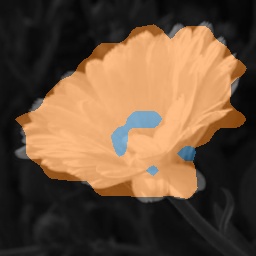} &
 \includegraphics[width=0.1\textwidth]{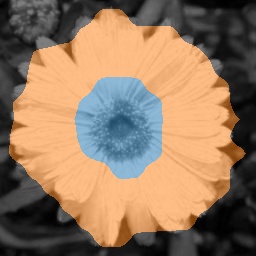}  &
 \includegraphics[width=0.1\textwidth]{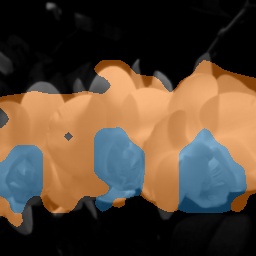}  &
 \includegraphics[width=0.1\textwidth]{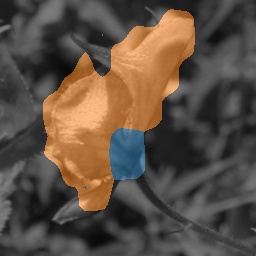}  &
 \includegraphics[width=0.1\textwidth]{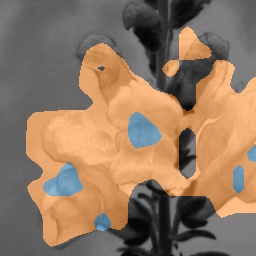}  &
 \includegraphics[width=0.1\textwidth]{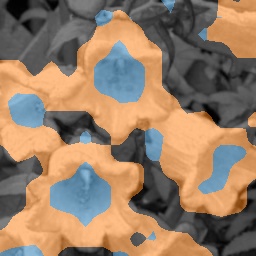}  &
 \includegraphics[width=0.1\textwidth]{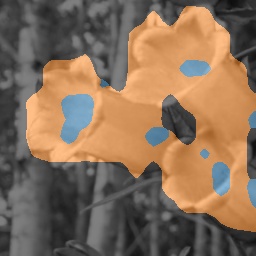}  &
 \includegraphics[width=0.1\textwidth]{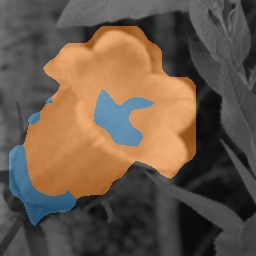}  &
 \includegraphics[width=0.1\textwidth]{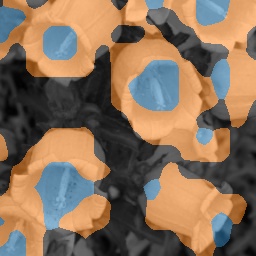} \\
\end{tabular}
\end{ssmall}
\caption{Qualitative results on Flowers for $K=2$.}
\label{fig:qual-flowers-supp}
\end{figure}
\begin{figure}[ht]
\centering
\setlength\tabcolsep{1.5pt} 
\begin{ssmall}
\centering
 \begin{tabular}{cccccccccc}
  \rotatebox[origin=tl]{90}{{\parbox{1.05cm}{\centering Image}}} &
 \includegraphics[width=0.1\textwidth]{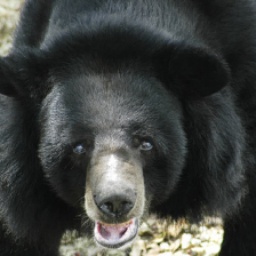} &
 \includegraphics[width=0.1\textwidth]{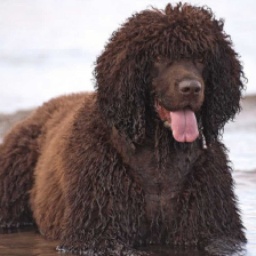}  &
 \includegraphics[width=0.1\textwidth]{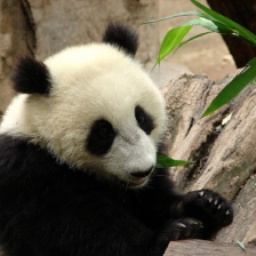}  &
 \includegraphics[width=0.1\textwidth]{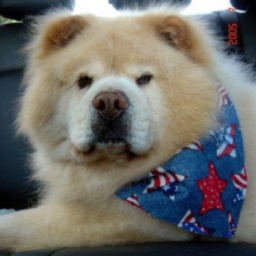}  &
 \includegraphics[width=0.1\textwidth]{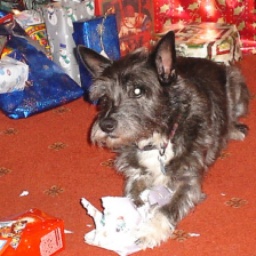}  &
 \includegraphics[width=0.1\textwidth]{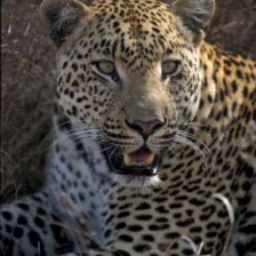}  &
 \includegraphics[width=0.1\textwidth]{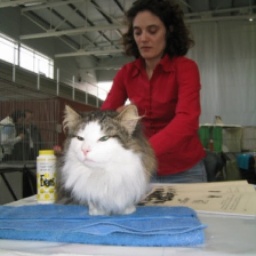}  &
 \includegraphics[width=0.1\textwidth]{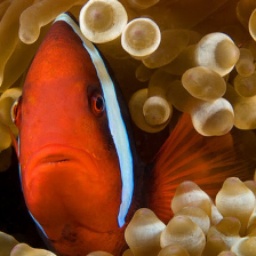}  &
 \includegraphics[width=0.1\textwidth]{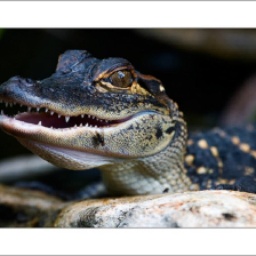}  \\
\rotatebox{90}{{\parbox{1.05cm}{\centering PDiscoNet\\+R101}}} &
     \includegraphics[width=0.1\textwidth]{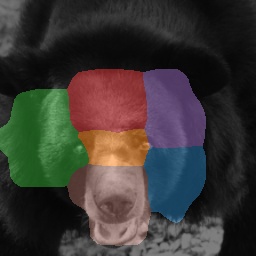} &
 \includegraphics[width=0.1\textwidth]{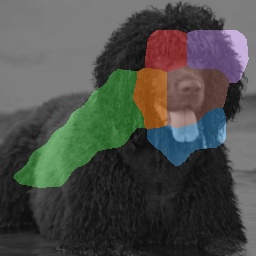}  &
 \includegraphics[width=0.1\textwidth]{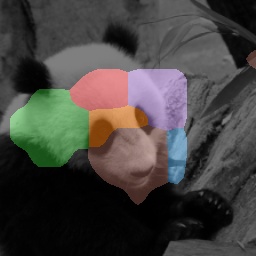}  &
 \includegraphics[width=0.1\textwidth]{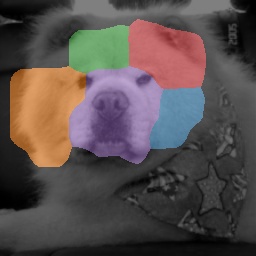}  &
 \includegraphics[width=0.1\textwidth]{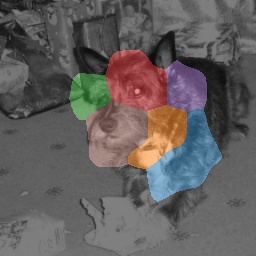}  &
 \includegraphics[width=0.1\textwidth]{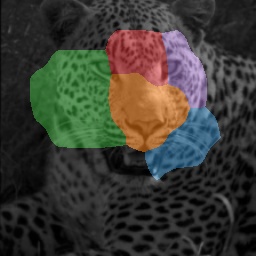}  &
 \includegraphics[width=0.1\textwidth]{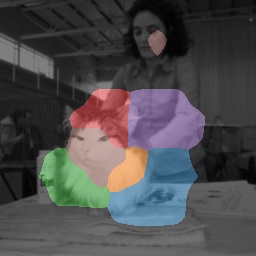}  &
 \includegraphics[width=0.1\textwidth]{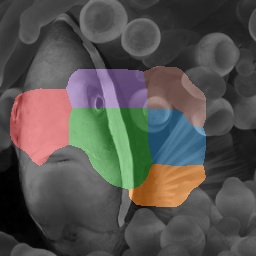}  &
 \includegraphics[width=0.1\textwidth]{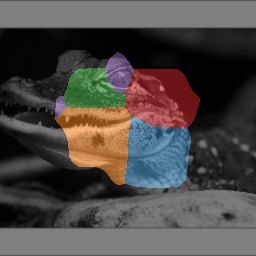} \\
 \rotatebox{90}{{\parbox{1.05cm}{\centering Ours \\+ R101}}} &
     \includegraphics[width=0.1\textwidth]{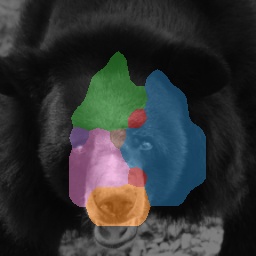} &
 \includegraphics[width=0.1\textwidth]{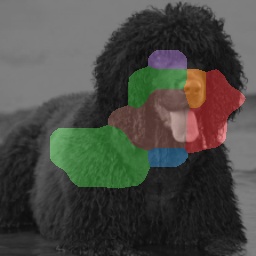}  &
 \includegraphics[width=0.1\textwidth]{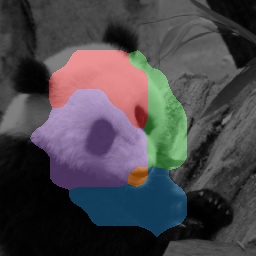}  &
 \includegraphics[width=0.1\textwidth]{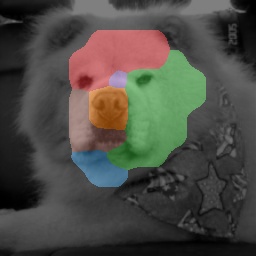}  &
 \includegraphics[width=0.1\textwidth]{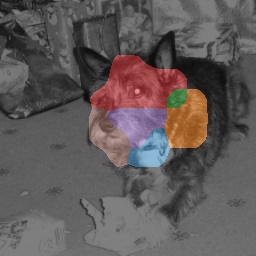}  &
 \includegraphics[width=0.1\textwidth]{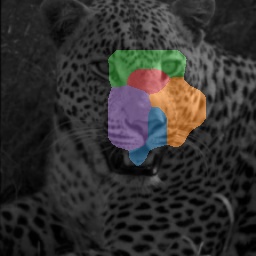}  &
 \includegraphics[width=0.1\textwidth]{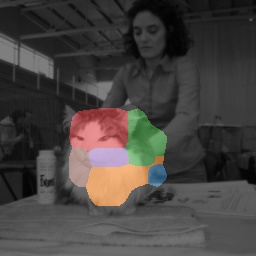}  &
 \includegraphics[width=0.1\textwidth]{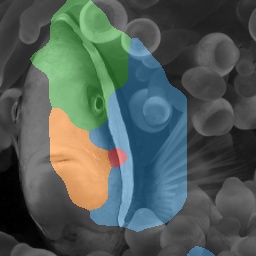}  &
 \includegraphics[width=0.1\textwidth]{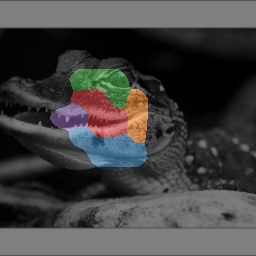} \\
 \rotatebox{90}{{\parbox{1.05cm}{\centering PDiscoNet + ViT-B}}} &
     \includegraphics[width=0.1\textwidth]{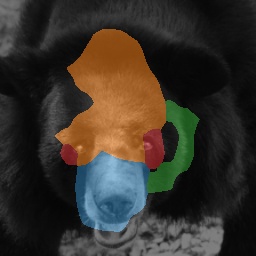} &
 \includegraphics[width=0.1\textwidth]{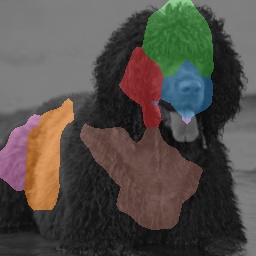}  &
 \includegraphics[width=0.1\textwidth]{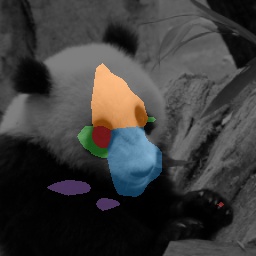}  &
 \includegraphics[width=0.1\textwidth]{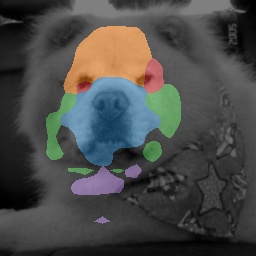}  &
 \includegraphics[width=0.1\textwidth]{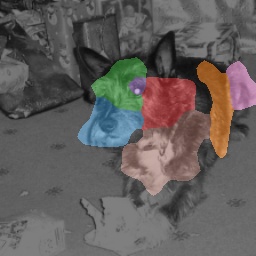}  &
 \includegraphics[width=0.1\textwidth]{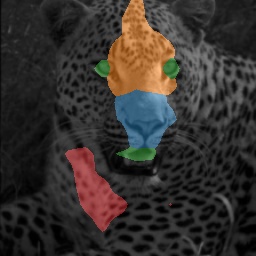}  &
 \includegraphics[width=0.1\textwidth]{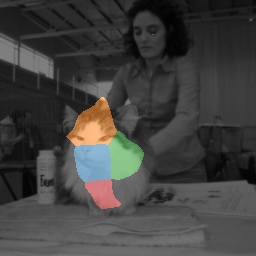}  &
 \includegraphics[width=0.1\textwidth]{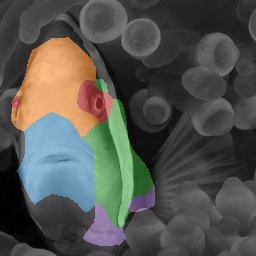}  &
 \includegraphics[width=0.1\textwidth]{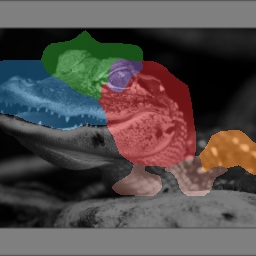} \\
 \rotatebox{90}{{\parbox{1.05cm}{\centering Ours \\+ ViT-B (frozen)}}} &
     \includegraphics[width=0.1\textwidth]{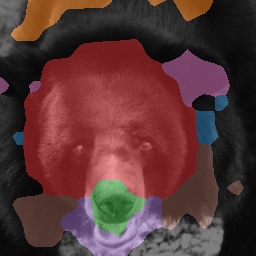} &
 \includegraphics[width=0.1\textwidth]{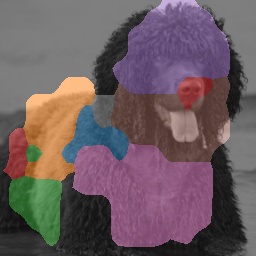}  &
 \includegraphics[width=0.1\textwidth]{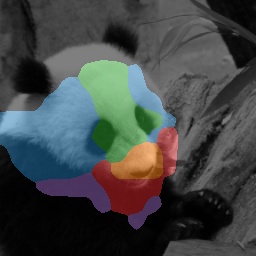}  &
 \includegraphics[width=0.1\textwidth]{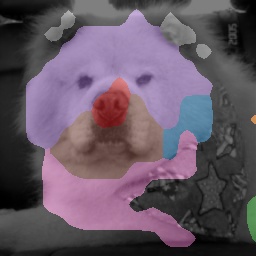}  &
 \includegraphics[width=0.1\textwidth]{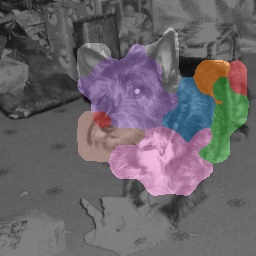}  &
 \includegraphics[width=0.1\textwidth]{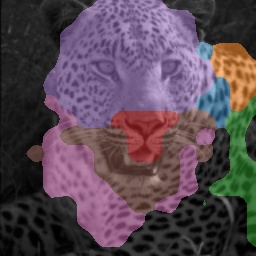}  &
 \includegraphics[width=0.1\textwidth]{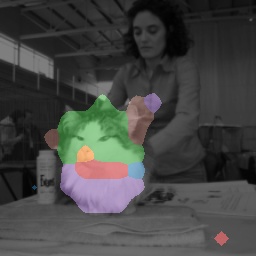}  &
 \includegraphics[width=0.1\textwidth]{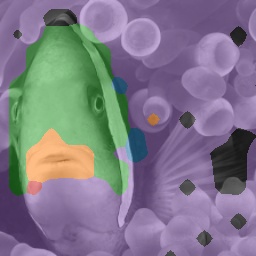}  &
 \includegraphics[width=0.1\textwidth]{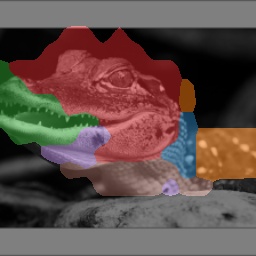} \\
\rotatebox[origin=tl]{90}{{\parbox{0.95cm}{\centering Ours}}} &
   \includegraphics[width=0.1\textwidth]{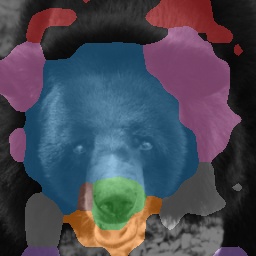} &
 \includegraphics[width=0.1\textwidth]{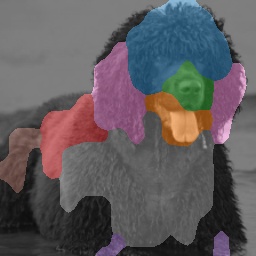}  &
 \includegraphics[width=0.1\textwidth]{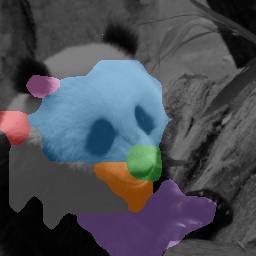}  &
 \includegraphics[width=0.1\textwidth]{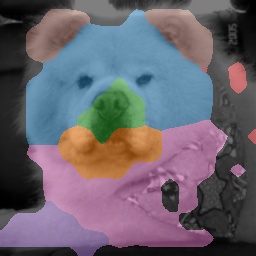}  &
 \includegraphics[width=0.1\textwidth]{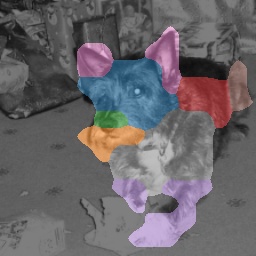}  &
 \includegraphics[width=0.1\textwidth]{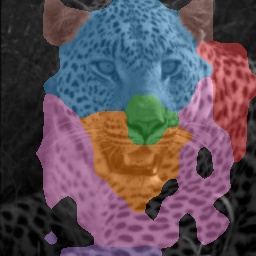}  &
 \includegraphics[width=0.1\textwidth]{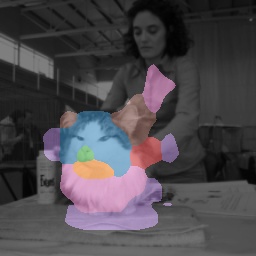}  &
 \includegraphics[width=0.1\textwidth]{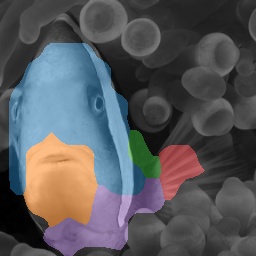}  &
 \includegraphics[width=0.1\textwidth]{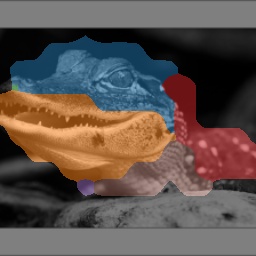} \\
\end{tabular}
\end{ssmall}
\caption{Qualitative results on PartImageNet OOD for $K=8$.}
\label{fig:qual-pimagenet-supp}
\end{figure}

In this experiment, we investigate the impact of different deep neural network backbones on the performance of our method. The results are summarized in~\cref{tab:sup-results}.

\noindent \textbf{Effect of Backbone Pre-Training for Part Discovery.}
From~\cref{tab:sup-results}, it is evident that our proposed part discovery priors perform optimally on the self-supervised DinoV2 ViT backbone. However, when compared to the PDiscoNet model with the same ResNet backbone, which utilizes the stricter concentration loss prior~\cite{van2023pdisconet, hung:CVPR:2019}, the resultant PDiscoFormer+R101 model exhibits inferior performance for both part discovery and classification. For instance, on the Oxford Flowers dataset, the PDiscoFormer+R101 model's training collapses for all tested values of $K$, resulting in classification accuracies of 8.4\%, 7.1\%, and 5\%, respectively. In contrast, the PDiscoNet+R101 model achieves significantly higher accuracies of 77.5\%, 83.1\%, and 81\% for the same values of $K$. Despite relatively better performance on the CUB and PartImageNet OOD datasets, the PDiscoFormer+R101 model still lags behind the self-supervised ViT and related methods from the literature.
These results indicate that a strong part shape prior, such as the concentration loss, is indeed required to obtain consistent part maps for ImageNet-pretrained CNN backbones. Moreover, they highlight the crucial role of the strong inductive biases that the ViT model learns during the self-supervised pre-training stage, enabling the use of a more flexible geometric prior such as the total variation loss.

\noindent \textbf{Frozen vs partially fine-tuned ViT.}
We observe from \cref{tab:sup-results} that our method performs best when we fine-tune the position embeddings, class, and register tokens along with our additional layers while keeping the rest of the ViT frozen. Although our losses can still operate on a completely frozen ViT, as shown in~\cref{tab:sup-results}, the performance is generally a bit lower, particularly for higher values of $K$.
For instance, on the CUB dataset with 16 parts, our method with the partially fine-tuned ViT achieved 55.8\% ARI, 73.4\% NMI, and 88.7\% classification accuracy, compared to 50\% ARI, 69.5\% NMI, and 85.1\% classification accuracy for our method with the fully frozen backbone. Similarly, on the PartImageNet OOD dataset with 50 parts, our method with the partially fine-tuned ViT achieved 62.2\% ARI, 46.3\% NMI, and 91\% classification accuracy, compared to 57.9\% ARI, 44.5\% NMI, and 90.6\% classification accuracy for our method with the fully frozen backbone.
These results indicate that some level of fine-tuning, combined with our proposed training objective function, is beneficial for discovering consistent parts from the self-supervised ViT.

\noindent \textbf{Qualitative Analysis.}
For our qualitative analysis, we examine the results obtained for the CUB ($K=8$ parts), Oxford Flowers ($K=2$ parts), and PartImageNet OOD ($K=8$ parts) datasets, as depicted in~\cref{fig:qual-cub-supp},~\cref{fig:qual-flowers-supp}, and~\cref{fig:qual-pimagenet-supp}, respectively. In~\cref{fig:qual-cub-supp}, our model with the partially fine-tuned self-supervised ViT (last row) demonstrates the most consistent results for part discovery and exhibits superior segmentation of discovered parts, such as the bird wings. The frozen ViT model (second last row) generally identifies consistent parts but occasionally misassigns background regions, such as the sky, as foreground parts. Notably, models with the self-supervised ViT backbone tend to more accurately segment foreground image regions and bird wings, indicating the effectiveness of representations learned during self-supervised pre-training for part discovery.
Turning to~\cref{fig:qual-flowers-supp}, only our models with the partially fine-tuned and frozen ViT (last two rows) successfully identify consistent parts. Among these, the partially fine-tuned model achieves better segmentation of the flowers and produces more semantically interpretable part assignments, with one part clearly corresponding to the flower calyx and the other to the corolla.
Finally, in~\cref{fig:qual-pimagenet-supp}, our model with the partially fine-tuned ViT (last row) demonstrates the most consistent and semantically interpretable parts. For instance, the blue part corresponds to the upper part of the animal's face, the green part to the snout, and the orange part to the mouth. The frozen ViT model (second last row) generally localizes the object of interest well but produces parts that are qualitatively more challenging to interpret. For example, it assigns the same part (red color) to the face of a bear and an alligator, although this part corresponds to the snout for dogs and leopards. Additionally, the PDiscoNet+ViT-B model consistently discovers the mouth of the animal as a part (blue color) and assigns the upper part of the face to the same part (orange color) in most cases. However, some of the other discovered parts (red and green colors) are less consistent and harder to interpret. Furthermore, this model achieves lower-quality segmentation of foreground parts due to the compactness prior of the concentration loss. In comparison, the PDiscoNet model and our model with the ResNet backbone find parts that are even more inconsistent, making them harder to interpret. Additionally, these models demonstrate poorer segmentation of the salient object in the image compared to our models with the frozen and partially fine-tuned ViT.

\section{Entropy Analysis of Part Attention Maps}
\label{sec-app:entropy}
\begin{figure}[t]
\centering
\begin{subfigure}[t]{0.495\textwidth}
\centering
\includegraphics[width=\textwidth]{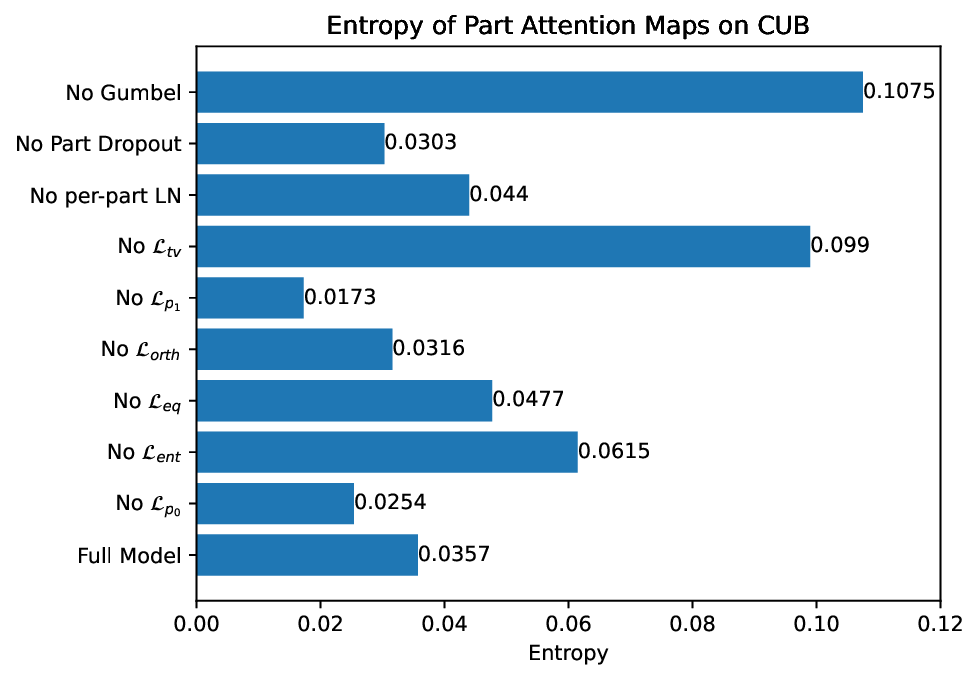}
\end{subfigure}
\hfill
\begin{subfigure}[t]{0.495\textwidth}
\centering
\includegraphics[width=\textwidth]{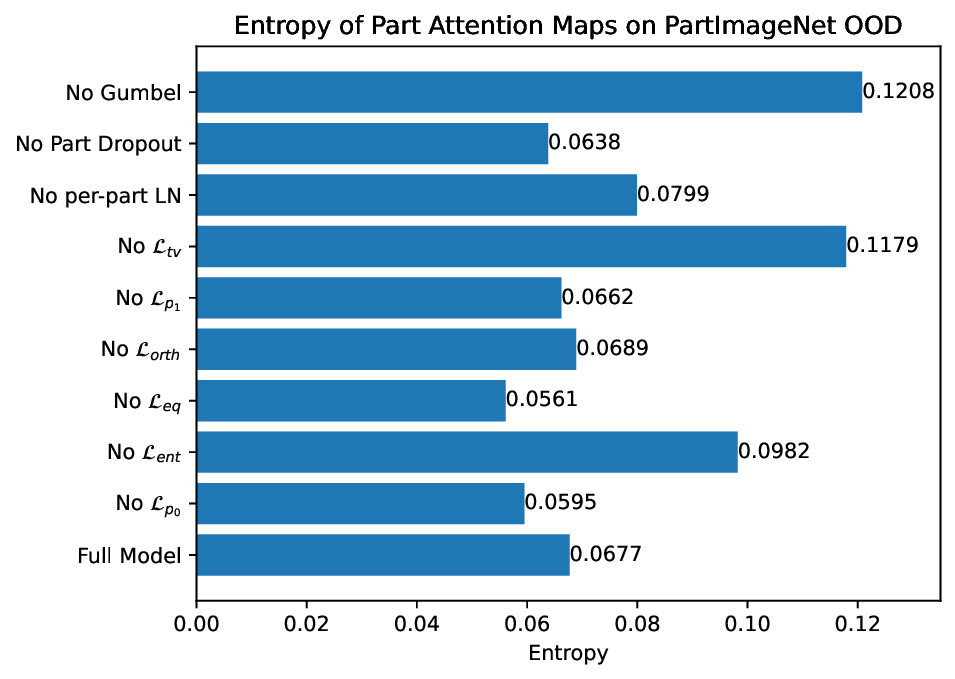}
\end{subfigure}
\caption{Entropy of part attention maps of the different model ablations on the CUB ($K=16$) and PartImageNet OOD ($K=25$) test sets.}
\label{fig:entropy}
\end{figure}

We conducted an analysis of the average entropy (see~\cref{eq:entropy_loss} in the main paper) in the part attention maps of our model ablations in the CUB ($K=16$) and PartImageNet OOD ($K=25$) datasets. The results are summarized in \cref{fig:entropy}.
From \cref{fig:entropy}, it is evident that the Gumbel-Softmax mechanism~\cite{jang2017categorical}, followed by the total variation loss ($\mathcal{L}_{tv}$) and entropy loss ($\mathcal{L}_{ent}$), significantly contribute to reducing the overall entropy in our part attention maps for both the CUB and PartImageNet OOD datasets. This underscores their crucial role in minimizing ambiguity in our part assignments and highlights the necessity of considering multiple components for effective entropy reduction, rather than solely focusing on minimizing the entropy loss ($\mathcal{L}_{ent}$).
By reducing information leakage between background and foreground assignments, lower entropy could help prevent the model from learning spurious correlations from background image regions~\cite{beery2018recognition,xiao2020noise,aniraj2023masking}, thereby enhancing model robustness across diverse imaging environments. Additionally, lower entropy between foreground part assignments would ensure that each discovered foreground part is a unique, independent feature. We believe this mechanism allows our model, as demonstrated in our empirical results, to effectively scale for different values of $K$ without any hyper-parameter tuning.

\section{Experiment on the PartImageNet Seg Dataset}
\label{sec-app:partimagenet-seg}
\begin{table}[t]
\centering
\caption{Quantitative results on the PartImageNet Seg dataset}
\label{tab:partimagenet-seg}
\begin{tabular}{c|cccc}
 & \multicolumn{4}{c}{PartImageNet Seg  (\%)}      \\ \cline{2-5} 
Method &
  \multicolumn{1}{c|}{K} &
  NMI$\uparrow$ &
  ARI$\uparrow$ &
  \begin{tabular}[c]{@{}c@{}}Top-1\\ Acc.$\uparrow$\end{tabular} \\ \hline
\multirow{5}{*}{PDiscoNet~\cite{van2023pdisconet}} &
  \multicolumn{1}{c|}{8} &
  10.69 &
  32.43 &
  85.03 \\
 & \multicolumn{1}{c|}{16} & 16.91 & 37.41 & 84.49 \\
 & \multicolumn{1}{c|}{25} & 18.44 & 40.70 & 84.12 \\
 & \multicolumn{1}{c|}{41} & 23.99 & 40.27 & 84.28 \\
 & \multicolumn{1}{c|}{50} & 21.98 & 39.40 & 84.32 \\ \hline
\multirow{5}{*}{\begin{tabular}[c]{@{}c@{}}PDiscoNet \\ + ViT-B\end{tabular}} &
  \multicolumn{1}{c|}{8} &
  16.68 &
  32.01 &
  88.36 \\
 & \multicolumn{1}{c|}{16} & 26.24 & 48.18 & 88.81 \\
 & \multicolumn{1}{c|}{25} & 25.93 & 46.81 & \textbf{88.90} \\
 & \multicolumn{1}{c|}{41} & 27.93 & 39.96 & 87.86 \\
 & \multicolumn{1}{c|}{50} & 28.68 & 34.13 & 87.73 \\ \hline
\multirow{5}{*}{\begin{tabular}[c]{@{}c@{}}PDiscoFormer\\ (Ours)\end{tabular}} &
  \multicolumn{1}{c|}{8} &
  20.29 &
  38.90 &
  87.65 \\
 & \multicolumn{1}{c|}{16} & 39.07 & 56.95 & 87.98 \\
 & \multicolumn{1}{c|}{25} & 43.18 & \textbf{64.61} & 88.15 \\
 & \multicolumn{1}{c|}{41} & \textbf{44.85} & 59.95 & 88.57 \\
 & \multicolumn{1}{c|}{50} & 44.06 & 60.10 & 88.65
\end{tabular}
\end{table}

We conducted an experiment on the \texttt{PartImageNet Seg} dataset, a recent addition to the PartImageNet family specifically tailored for image classification tasks~\cite{he2022partimagenet}. This dataset comprises 158 classes organized into 11 super-classes, with a total of 41 part classes, maintaining consistency with the \texttt{PartImageNet OOD} version. With a training set comprising 21,662 images and a test set containing 2,405 images, assessing the part discovery and classification capabilities of our method on this dataset can provide further insights into its overall generalization. Given the absence of comparable results in the existing literature for unsupervised part discovery on this new dataset, we conducted our experiments from scratch for all models presented in~\cref{tab:partimagenet-seg}. This approach ensures that our evaluation is thorough and offers valuable insights into the effectiveness of our method in this evaluation setting.

\noindent \textbf{Quantitative Results.}
\label{subsec:sup-pimagenet-quant}
We observe from~\cref{tab:partimagenet-seg} that our method with the self-supervised ViT backbone consistently outperforms the state-of-the-art method in terms of consistency in part discovery, as indicated by the NMI and ARI scores. For instance, for $K=41$ parts, our method achieves NMI and ARI scores of 44.9\% and 60\%, respectively, compared to 27.9\% and 40\% by the PDiscoNet+ViT-B method and 24\% and 40.3\% by the original PDiscoNet. Additionally, similar to our previous experiments, we observe that the classification accuracy for the PDiscoNet method (also with the self-supervised ViT backbone) reduces with an increase in the number of parts to be discovered ($K$). Specifically, from~\cref{tab:partimagenet-seg}, our model's classification accuracy increases from 87.7\% to 88.7\%, while it reduces from 88.4\% to 87.7\% for PDiscoNet+ViT-B and from 85\% to 84.3\% for the original PDiscoNet.
These results further indicate the generalization capability of our model for datasets containing objects with diverse part shapes.

\begin{figure}[t]
\centering
\setlength\tabcolsep{1.5pt} 
\begin{ssmall}
\centering
 \begin{tabular}{cccccccccc}
 \rotatebox[origin=tl]{90}{{\parbox{1.05cm}{\centering Image}}} &
 \includegraphics[width=0.1\textwidth]{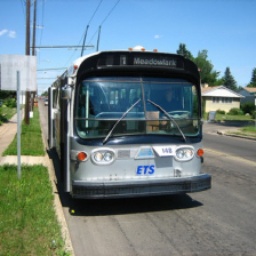} &
 \includegraphics[width=0.1\textwidth]{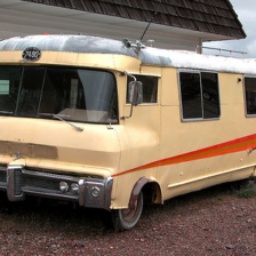}  &
 \includegraphics[width=0.1\textwidth]{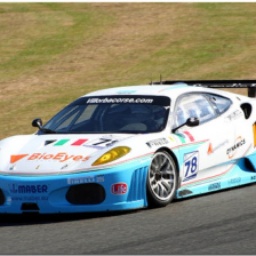}  &
 \includegraphics[width=0.1\textwidth]{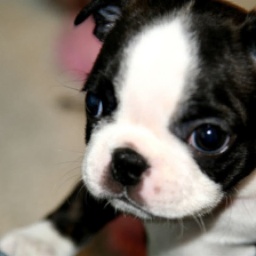}  &
 \includegraphics[width=0.1\textwidth]{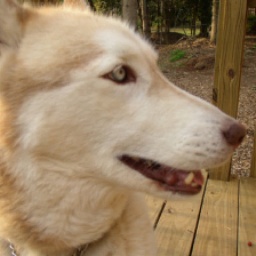}  &
 \includegraphics[width=0.1\textwidth]{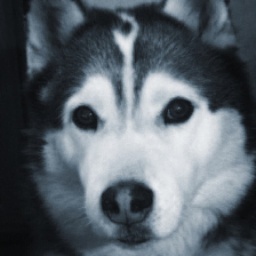}  &
 \includegraphics[width=0.1\textwidth]{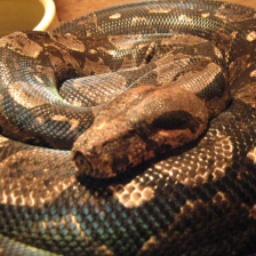}  &
 \includegraphics[width=0.1\textwidth]{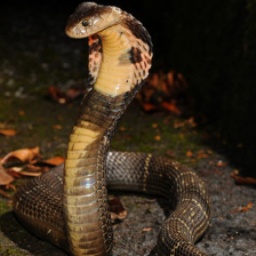}  &
 \includegraphics[width=0.1\textwidth]{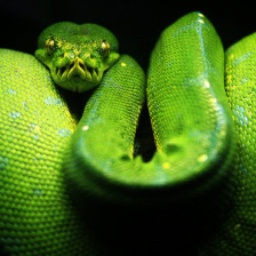}  \\
\rotatebox{90}{{\parbox{1.05cm}{\centering PDiscoNet}}} &
 \includegraphics[width=0.1\textwidth]{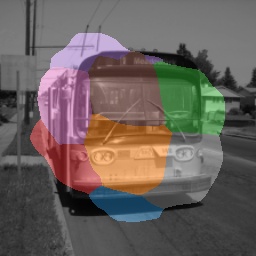} &
 \includegraphics[width=0.1\textwidth]{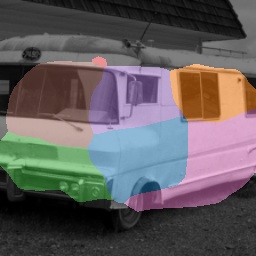}  &
 \includegraphics[width=0.1\textwidth]{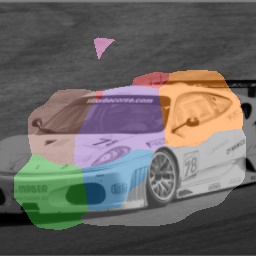}  &
 \includegraphics[width=0.1\textwidth]{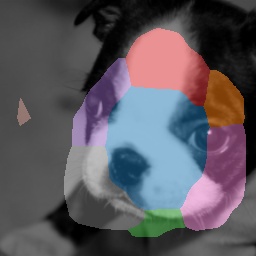}  &
 \includegraphics[width=0.1\textwidth]{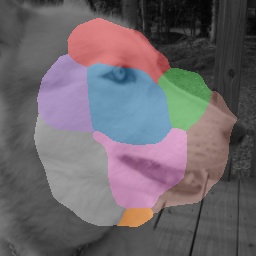}  &
 \includegraphics[width=0.1\textwidth]{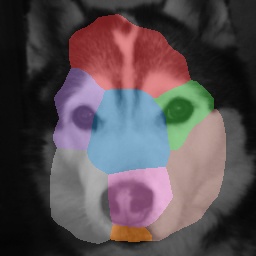}  &
 \includegraphics[width=0.1\textwidth]{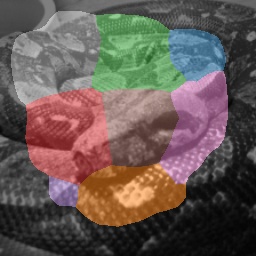}  &
 \includegraphics[width=0.1\textwidth]{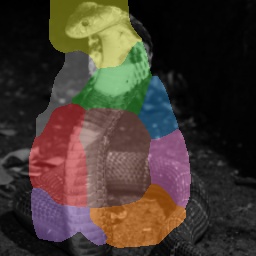}  &
 \includegraphics[width=0.1\textwidth]{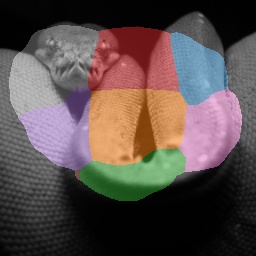} \\
 \rotatebox{90}{{\parbox{1.05cm}{\centering PDiscoNet + ViT-B}}} &
 \includegraphics[width=0.1\textwidth]{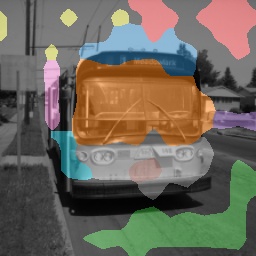} &
 \includegraphics[width=0.1\textwidth]{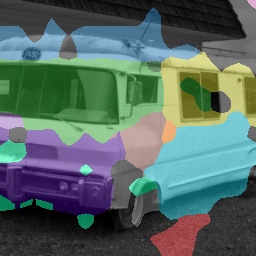}  &
 \includegraphics[width=0.1\textwidth]{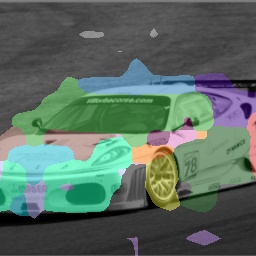}  &
 \includegraphics[width=0.1\textwidth]{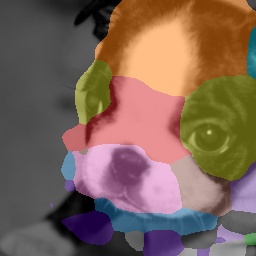}  &
 \includegraphics[width=0.1\textwidth]{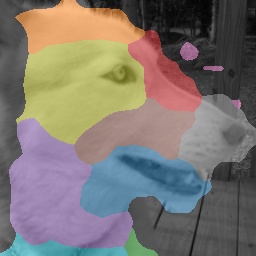}  &
 \includegraphics[width=0.1\textwidth]{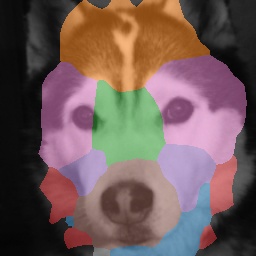}  &
 \includegraphics[width=0.1\textwidth]{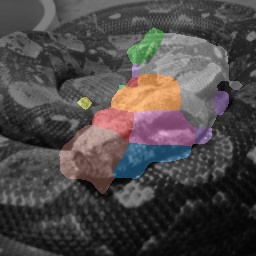}  &
 \includegraphics[width=0.1\textwidth]{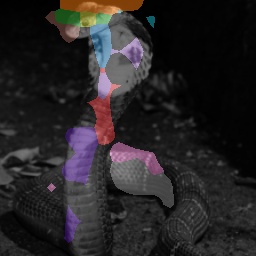}  &
 \includegraphics[width=0.1\textwidth]{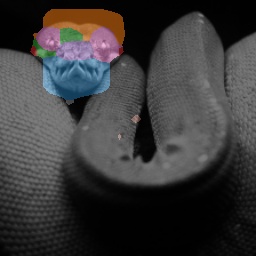} \\
\rotatebox[origin=tl]{90}{{\parbox{0.95cm}{\centering Ours}}} &
\includegraphics[width=0.1\textwidth]{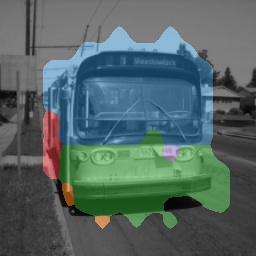} &
 \includegraphics[width=0.1\textwidth]{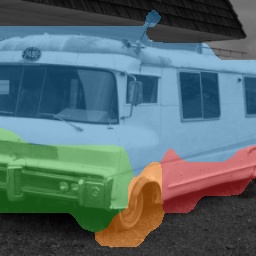}  &
 \includegraphics[width=0.1\textwidth]{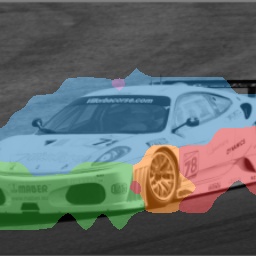}  &
 \includegraphics[width=0.1\textwidth]{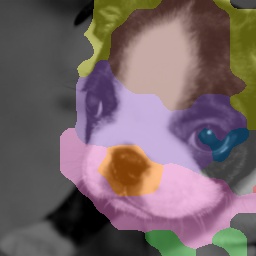}  &
 \includegraphics[width=0.1\textwidth]{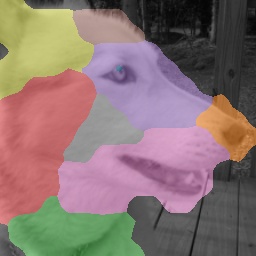}  &
 \includegraphics[width=0.1\textwidth]{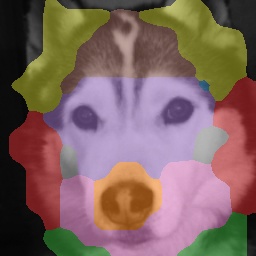}  &
 \includegraphics[width=0.1\textwidth]{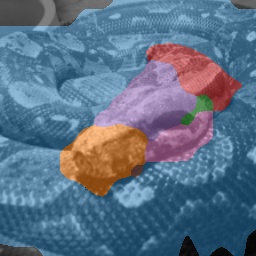}  &
 \includegraphics[width=0.1\textwidth]{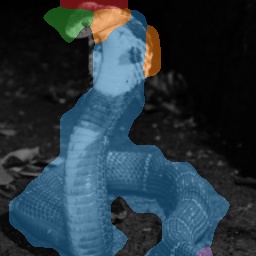}  &
 \includegraphics[width=0.1\textwidth]{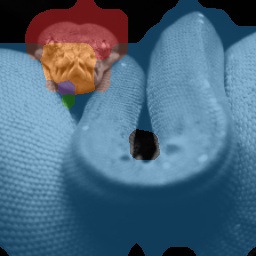} \\
\end{tabular}
\end{ssmall}
\caption{Qualitative results on PartImageNet Seg for $K=25$ parts. The first 3 images belong to the super-class \textit{Car}, the next 3 to the super-class \textit{Quadruped} and the final 3 images belong to the super-class \textit{Snake}.}
\label{fig:qual-pimagenetseg}
\end{figure}
\noindent \textbf{Qualitative Results.}
Qualitative results for the PartImageNet Seg dataset with $K=25$ parts are shown in~\cref{fig:qual-pimagenetseg}, featuring three images each for the super-classes \textit{Car}, \textit{Quadruped}, and \textit{Snake}.
From~\cref{fig:qual-pimagenetseg}, it is evident that our model is capable of consistently identifying parts with diverse shapes, such as the vehicle bumper, the dog's snout and ears, and the snake's body. The parts discovered by our model are also, on average, easier to interpret semantically. In contrast, the compared methods struggle to localize parts with irregular shapes, such as the snake body, and are generally less consistent, making them harder to interpret.
\end{document}